\pdfoutput=1
\documentclass[journal]{IEEEtran}

\usepackage{multirow}
\usepackage{bm}
\usepackage{color}
\usepackage{epsfig}
\usepackage{graphicx}
\usepackage{amsmath}
\usepackage{amssymb}
\usepackage{algorithm} 
\usepackage{algorithmic} 

\usepackage{amsthm}
\usepackage{array}
\usepackage{colortbl}
\usepackage{bbding}
\usepackage{url}
\usepackage{hyperref}
\usepackage{booktabs}
\usepackage{multirow}
\usepackage{cite}
\usepackage{graphicx}
\usepackage{tabularx}
\usepackage{epstopdf}
\usepackage{subfigure}
\theoremstyle{remark}

\newcommand{\eg}{\textit{e.g.,~}}
\newcommand{\ie}{\textit{i.e.,~}}

\newcommand{\etal}{\textit{et al.~}}

\begin{document}
	\title{Convolutional Fine-Grained Classification with Self-Supervised Target Relation Regularization}
 
\author{Kangjun Liu, Ke Chen,~\IEEEmembership{Member,~IEEE,}~and~Kui Jia,~\IEEEmembership{Member,~IEEE}
\IEEEcompsocitemizethanks{
\IEEEcompsocthanksitem This work is supported in part by the National Natural Science Foundation of China (Grant No.: 61902131), the Program for Guangdong Introducing Innovative and Enterpreneurial Teams (Grant No.: 2017ZT07X183), and the Guangdong Provincial Key Laboratory of Human Digital Twin (Grant No.: 2022B1212010004). K. Chen is the corresponding author of this work.
\IEEEcompsocthanksitem 
K. Liu, K. Chen and K. Jia are with the South China University of Technology, Guangzhou 510641, China, and also with the Peng Cheng Laboratory, Shenzhen, China. 
E-mails: wikangj.liu@mail.scut.edu.cn; chenk@scut.edu.cn; kuijia@scut.edu.cn.}
}
	
\markboth{Journal of \LaTeX\ Class Files,~Vol.~X, No.~X, June~2021}%
{Shell \MakeLowercase{\textit{et al.}}: Bare Demo of IEEEtran.cls for IEEE Journals}	

\maketitle

\begin{abstract}
Fine-grained visual classification can be addressed by deep representation learning under supervision of manually pre-defined targets (\eg one-hot or the Hadamard codes).
Such target coding schemes are less flexible to model inter-class correlation and are sensitive to sparse and imbalanced data distribution as well.
In light of this, this paper introduces a novel target coding scheme -- dynamic target relation graphs (DTRG),
which, as an auxiliary feature regularization, is a self-generated structural output to be mapped from input images.
Specifically, online computation of class-level feature centers is designed to generate cross-category distance in the representation space, which can thus be depicted by a dynamic graph in a non-parametric manner. 
Explicitly minimizing intra-class feature variations anchored on those class-level centers can encourage learning of discriminative features. 
Moreover, owing to exploiting inter-class dependency, the proposed target graphs can alleviate data sparsity and imbalanceness in representation learning. 
Inspired by recent success of the mixup style data augmentation, 
this paper introduces randomness into soft construction of dynamic target relation graphs to further explore relation diversity of target classes. 
Experimental results can demonstrate the effectiveness of our method on a number of diverse benchmarks of multiple visual classification tasks, especially achieving the state-of-the-art performance on popular fine-grained object benchmarks and superior robustness against sparse and imbalanced data.
Source codes are made publicly available at \href{https://github.com/AkonLau/DTRG}{https://github.com/AkonLau/DTRG}.
\end{abstract}




\IEEEpeerreviewmaketitle

\section{Introduction}\label{sec:introduction}

\IEEEPARstart{T}{he} problem of image classification \cite{Krizhevsky2012ImageNetCW, He2016DeepRL, Huang2017DenselyCC} aims to categorize still images into semantic classes according to visual appearance of objects in images. 
Such a problem is made more challenging to recognize fine-grained objects, \ie subclass of animal species such as birds \cite{wah2011caltech} and dogs \cite{khosla2011novel}, or the brand and model of man-made products such as cars \cite{krause20133d} and aircraft \cite{maji2013fine}, which can be addressed by deep representation learning \cite{Reed2016LearningDR,Chen2018FineGrainedRL} (consists of a deep feature encoder and a typical classifier).
The challenges of classifying fine-grained classes in deep feature encoding lie in coping with cross-class visual similarity and within-class variations intrinsically caused by object poses, in addition to other extrinsic imaging factors such as varying viewing angles and illumination.

Most of existing methods \cite{Xie2013HierarchicalPM, zhang2014part, branson2014bird, girshick2014rich, shih2015part, huang2016part, zheng2019learning, fu2017look, lin2015bilinear, yu2018hierarchical, zhuang2020learning, huang2021snapmix, sun2018multi} focus on learning discriminative features to discover subtle appearance difference spatially distributed on local parts. 
A number of existing methods \cite{Xie2013HierarchicalPM, zhang2014part, branson2014bird, girshick2014rich, shih2015part} attempt to locate and align pivotal parts of object by leveraging extra supervision of key points or bounding boxes of parts, which demand laborious annotations and thus lead to the limited size of training data.
More recent part-based methods \cite{huang2016part, zheng2019learning} and global feature encoding \cite{lin2015bilinear, yu2018hierarchical, zhuang2020learning, huang2021snapmix, cai2019convolutional, Ding2021APCNNWS} can either discover distinguishable regions in a weakly-supervised learning manner, or enrich global representations with high-order information or instance-wise discrimination,  to avoid utilizing extra part annotations.
\begin{figure}[t]
    \centering
    \subfigure{ 
    \includegraphics[width=\linewidth]{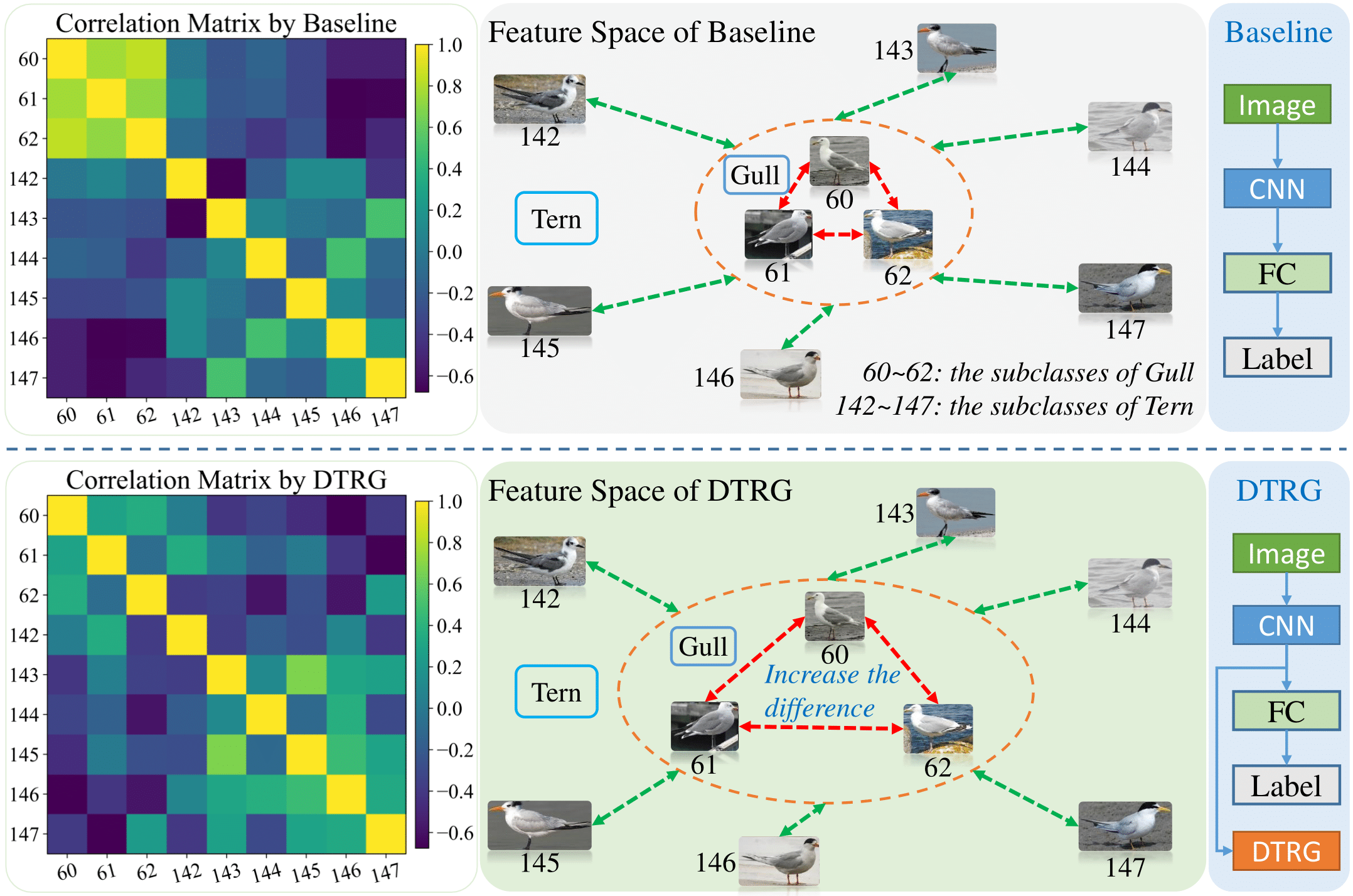}}\vspace{-0.2cm}
    \subfigure{ 
    \resizebox{0.95\linewidth}{!}{    
    \begin{tabular}{@{}lcccccccccc@{}}
    \toprule
    & \multicolumn{9}{c}{Selected Class ID of CUB (\%)}                                                                                             & \multicolumn{1}{l}{}    \\ \cmidrule(lr){2-10}
    Method  & 60            & 61           & 62            & 142           & 143           & 144           & 145           & 146           & 147           & \multicolumn{1}{l}{AVG} \\ \midrule
    Baseline     & 62.1          & 96.7         & 53.3          & 56.7          & 60.0          & \textbf{43.3} & 70.0          & \textbf{60.0} & 86.7          & 65.4                    \\
    DTRG  & \textbf{65.5} & \textbf{100} & \textbf{63.3} & \textbf{73.3} & \textbf{73.3} & 40.0          & \textbf{76.7} & 56.7          & \textbf{93.3} & \textbf{71.4}           \\ \bottomrule
    \end{tabular}}    
    }

\caption{Comparative evaluation with the vanilla CNN baseline (top) and the proposed dynamic target relation graph (DTRG) regularized CNN model (middle), where the backbone of both models is identical, on the CUB-200-2011 benchmark \cite{wah2011caltech}. 
For illustrative purpose, a subset of fine-grained bird breed classes are selected from the CUB-200-2011, where the id of object classes are shown below the images.
The classes within and outside the red dashed ellipses are from the parent classes of Gulls and Terns, respectively.
\textcolor{black}{The selected per-class accuracy is shown in the bottom.}
Evidently, class-level representations of the visually similar classes (\ie three Gull classes within the red ellipses) with the proposed DTRG regularization can be more distinguishable than the vanilla model, which can also be observed from the confusion matrices on the left handside, owing to the introduction of latent correlation across target classes. }
\label{fig:introduce}
\vspace{-0.2cm}
\end{figure}

The aforementioned methods are typically under the supervision of one-hot target code \cite{He2016DeepRL, Huang2017DenselyCC} or its soft variants such as label smoothing \cite{Szegedy2016RethinkingTI, zhang2020delving}.
Very few works including the label relation graph \cite{Deng2014LargeScaleOC} and the Hadamard code \cite{Yang2015DeepRL} attempt to exploit inter-class correlation to avoid inconsistent predictions and also transfer knowledge across classes. 
On one hand, the label relation graph can enrich the structure in the target space based on prior knowledge, which desires additional expensive professional efforts on annotating target relation.
On the other hand, the target codes based on the Hadamard code aim to implicitly model inter-class relation using its redundant positive elements, whose optimization could be sub-optimal due to random assignment of target codes to semantic classes.
More importantly, existing target coding schemes are manually pre-defined and thus less flexible to model complex relations between targets. 

In light of this, a novel target relation scheme is proposed as an auxiliary feature regularization in a deep representation learning framework, which can discover latent correlation in the target space with no price of additional human efforts.
Moreover, owing to mining target correlation, the proposed method can inherently  mitigate negative effects of sparse and imbalanced data distribution, as samples from correlated classes can support to distinguish those from long-tailed classes.
To this end, two key issues need to be addressed, \ie discovering class-level representation within each fine-grained category and constructing target relation representation across classes.

To generate class-level centers in the feature space, existing parametric center generation method \cite{wen2016discriminative} learns to discover the class-level centers by constraining distance between samples' features and their corresponding centers.
Consequently, such a learnable parametric center generation can minimize the intra-class variance, but is not stable and robust against sparse and imbalanced data distribution.
Such an observation inspires us to propose an online center generation method in a parameter-free fashion.
We observe that for class-level representations (\ie feature centers), which are the output of deep feature encoding, they are non-consistent when training epoch evolves.
In this way, generation of dynamic class-level representations in our scheme is extreme simple, which are iteratively updated by the mean value of all samples' features from the same class in the last epoch.
Encouraged by success of the center loss in \cite{wen2016discriminative}, we adaptively propose an online center loss to minimize intra-class feature variations to improve discrimination of features.
The proposed online center loss shares similar concept with the online label smoothing \cite{zhang2020delving}, but their goal is to generate soft labels revealing data distribution rather than capture inter-class target relation.

Given the class-level representations, we propose a novel dynamic target relation graph (DTRG) via measuring relative distance in the feature space to regularize deep feature encoding to incorporate latent inter-class relation as an extra similarity metric of feature encoding.
As a result, samples from each class are not only supervised by the corresponding one-hot target code, but also constrained by all the other class-level centers in the feature space, as shown in Fig. \ref{fig:introduce}.
Moreover, exploiting target relation in our scheme can also alleviate sparse and long tailed data distribution, which can further improve robustness of deep representations.
Note that, the DTRG is non-parametric and can be readily inserted in any existing classification backbones for enhancing representation learning.
To further expand the rich structure in the target space, recent soft-label based data augmentation such as vanilla Mixup\cite{zhang2018mixup}, CutMix \cite{Yun2019CutMixRS} and more recent SnapMix\cite{huang2021snapmix} can be employed.
Extensive experiments on diverse public benchmarks can verify the effectiveness of our method, especially for a long-tail distributed data. 

Main contributions can be summarized as follows:
\begin{enumerate}
    \item We propose a novel non-parametric feature regularization to discover latent correlation across target classes in deep representation learning for visual classification, {which is generic and can be readily adopted in existing classification methods.}
    \item Technically, we develop an online center loss to generate dynamic class-level representations, which are then utilized to construct a novel dynamic target relation graph. The proposed scheme can not only minimize intra-class feature variance but also capture inter-class correlation to encourage discriminative features.
    \item Diverse inter-class target relation in the proposed DTRG can be further explored via semantic interpolation, as the mixup-style data augmentation.    
    \item Our method can beat recent comparative methods on popular benchmarks of multiple visual classification tasks, achieving the state-of-the-art accuracy on fine-grained object benchmarks and superior robustness against sparse and imbalanced data.
\end{enumerate}

The remainder of the paper is organized as follows. We briefly investigate related work in Sec. \ref{sec:relatedWorks}, and introduce our proposed graph similarity based feature regularization in a typical deep representation learning in Sec. \ref{sec: methods}. Experiments and evaluations on three widely-used fine-grained datasets and  other popular visual classification datasets are presented in Sec. \ref{sec: exps}, then followed by our conclusion in Sec. \ref{sec: conclusion}.

\section{Related Works}
\label{sec:relatedWorks}

Fine-grained visual classification has attracted wide attention in computer vision for decades, and a large number of algorithms have been proposed in the context of deep learning.
Existing methods can be categorized into two main groups -- part-based \cite{Xie2013HierarchicalPM, zhang2014part, shih2015part, huang2016part, branson2014bird, Jaderberg2015SpatialTN, Sermanet2015AttentionFF, Xiao2015TheAO, zheng2017learning, fu2017look,zheng2019learning, ding2019selective, zhang2021multi} and part-free \cite{Gao2016CompactBP, lin2015bilinear, yu2018hierarchical, Wang2015MultipleGD, Zhang2019LearningAM, Ding2021APCNNWS, Chang2020TheDI, Chang2020YourI,  zhang2016embedding, wang2017deep,sohn2016improved, sun2018multi, dubey2018pairwise, zhuang2020learning,chen2019destruction, zhou2020look, p2pnet2022}, which will be respectively investigated in details as follows.

\subsection{Part-based Methods} 
The pioneering methods \cite{Xie2013HierarchicalPM, zhang2014part, shih2015part, huang2016part, branson2014bird, chen2016learning, zhang2019part} attempted to learn discriminative features by detecting and aligning the key parts of objects. 
Zhang \etal \cite{zhang2014part} proposed a bottom-up region proposal based method for part location detection, which aims to learn a pose-normalized representation to discount negative effects of object poses. 
Huang \etal \cite{huang2016part} proposed a Part-stacked Network for locating object parts by modeling the subtle differences of object parts. 
Ge \etal \cite{Ge2019WeaklySC} built a complementary part detection model and then fused features of these different parts with a bi-direction LSTM network. 
Owing to its excellent performance for fine-grained features learning, the part-based methods \cite{zheng2019learning, Ge2019WeaklySC} remain active and recent concern is to spot discriminative parts without extra part annotations.
Moreover, a number of attention-based methods \cite{Jaderberg2015SpatialTN, Sermanet2015AttentionFF, Xiao2015TheAO, zheng2017learning, fu2017look, ding2019selective, zhang2021multi, Rao2021CounterfactualAL} are designed for detecting the discriminative regions and learning features by imitating the attention mechanism of humans. 
In \cite{fu2017look} and \cite{zhang2021multi}, an attention module was designed to detect discriminative regions iteratively, from which fine-grained local features are extracted for object classification.
\textcolor{black}{
Recently, inspired by the self-attention mechanism of vision transformer (ViT) \cite{Dosovitskiy2021AnII}, an amount of transformer-based algorithms have been proposed to extract discriminative features of fine-grained objects.
He \etal\cite{He2021TransFGAT} proposed a transformer-based method via selecting discriminative image patches with an attention map.
Wang \etal \cite{wang2021feature} proposed a feature fusion vision transformer (FFVT) to enrich representations of fine-grained classes via
aggregating local information across layers. 
}
Evidently, existing part-based methods aim to improve feature discrimination via enhancing distinguished details in local regions rather than exploiting inter-class relation as our method.

\subsection{Part-free Methods} 

An alternative group of fine-grained classifiers focus on global feature encoding, which can be further divided into two sub-categories -- fusion-based \cite{Gao2016CompactBP, lin2015bilinear, yu2018hierarchical, Wang2015MultipleGD, Zhang2019LearningAM, Ding2021APCNNWS, Chang2020TheDI, Chang2020YourI} and regularization-based \cite{chen2019destruction, zhou2020look, p2pnet2022}.

\vspace{0.1cm}\noindent
\textcolor{black}{
\textbf{Fusion-based --}
Lin \etal \cite{lin2015bilinear} were the first to introduce the bilinear model, which can learn discriminative features by using the outer product of two pooled features from different CNNs. 
Inspired by the bilinear pooling in \cite{lin2015bilinear}, there appears an amount of works \cite{Kong2017LowRankBP, Li2017IsSI, Li2018TowardsFT, Wang2018GlobalGM, Gou2018MoNetME, Zheng2019LearningDB, Yu2020TowardFA, Song2021WhyAM, Yu2021FastAC,Wang2021DeepCM, yu2022efficient, Koniusz2022PowerNI} about representation learning with construction of high-order features.
Gao \etal\cite{Gao2016CompactBP} proposed a compact bilinear pooling method to reduce feature dimensionality of high order representations, and more efficient compact bilinear pooling have been achieved by the Shifted Random Maclaurin \cite{Yu2021FastAC} and the Kronecker Product \cite{yu2022efficient}.
Li \etal\cite{Li2017IsSI} proposed a matrix power normalized covariance method to explore effective high-order statistics, and further improved the training of global covariance pooling via an iterative matrix square root normalization \cite{Li2018TowardsFT}, which inspires a number of follow-uppers \cite{Gou2018MoNetME, Song2021WhyAM, Koniusz2022PowerNI}.
Zheng \etal \cite{Zheng2019LearningDB} proposed a deep bilinear transformation block to produce and aggregate intra-group high-order features based on semantic grouping of feature channels.
Beyond exploiting intra-layer feature maps, Yu \etal \cite{yu2018hierarchical} designed a cross-layer bilinear pooling to enrich representations via fusing features from different granularities and spatial locations.
}
Different from the bilinear-based pooling, the methods in \cite{Wang2015MultipleGD, Zhang2019LearningAM} both proposed a cross-granularity framework for learning a mixture of features from different levels, while Du \etal \cite{du2020fine} proposed multi-granularity feature learning with a progressive training strategy. 

\vspace{0.1cm}\noindent\textbf{Regularization-based --} Alternatively, regularization-based methods \cite{chen2019destruction, zhou2020look, p2pnet2022} adopt prevalent self-supervised learning as regularization for enhancing discriminate features learning. 
A number of metric learning based methods \cite{zhang2016embedding, wang2017deep,sohn2016improved, sun2018multi, dubey2018pairwise, zhuang2020learning} were proposed to regularize feature encoding with similarity constraints.
Zhang \etal \cite{zhang2016embedding} adopted the squared Euclidean distance as similarity measurement between two $l_2$-normalized vectors and then generalized the triplet loss with a triplet network. 
As traditional metric learning methods always adopting distance measure, Wang \etal \cite{wang2017deep} firstly proposed an angular constraint for deep metric learning by encoding the angular relation among triplets, which  
can be high-order relation and thus able to reveal more local structure. 
Methods in \cite{sohn2016improved} and \cite{sun2018multi} both leveraged the multi-pair triplet loss to optimize the convergence, and Dubey \etal \cite{dubey2018pairwise} proposed a pairwise confusion regularization by minimizing the squared Euclidean distance between the output distribution of random pairs for preventing from over-fitting. 
Recently, Zhuang \etal \cite{zhuang2020learning} proposed an Attentive Pairwise Interaction architecture to learn the mutual feature between input pairs and then taking feature priorities into account with a score ranking loss.

The proposed method falls into the latter group of regularization based feature learning. 
However, different from aforementioned approaches, our dynamic target relation graph as self-generated constraint reveals latent inter-class correlation based on class-level representations, rather than traditional instance-level similarity measure, 
which can alleviate sparse and long tailed data distribution and therefore improve robustness of visual representations.

\section{Dynamic Target Relation Regularization}
\label{sec: methods}
\begin{figure*}[htbp]
    \centering
    \includegraphics[width=0.95\linewidth]{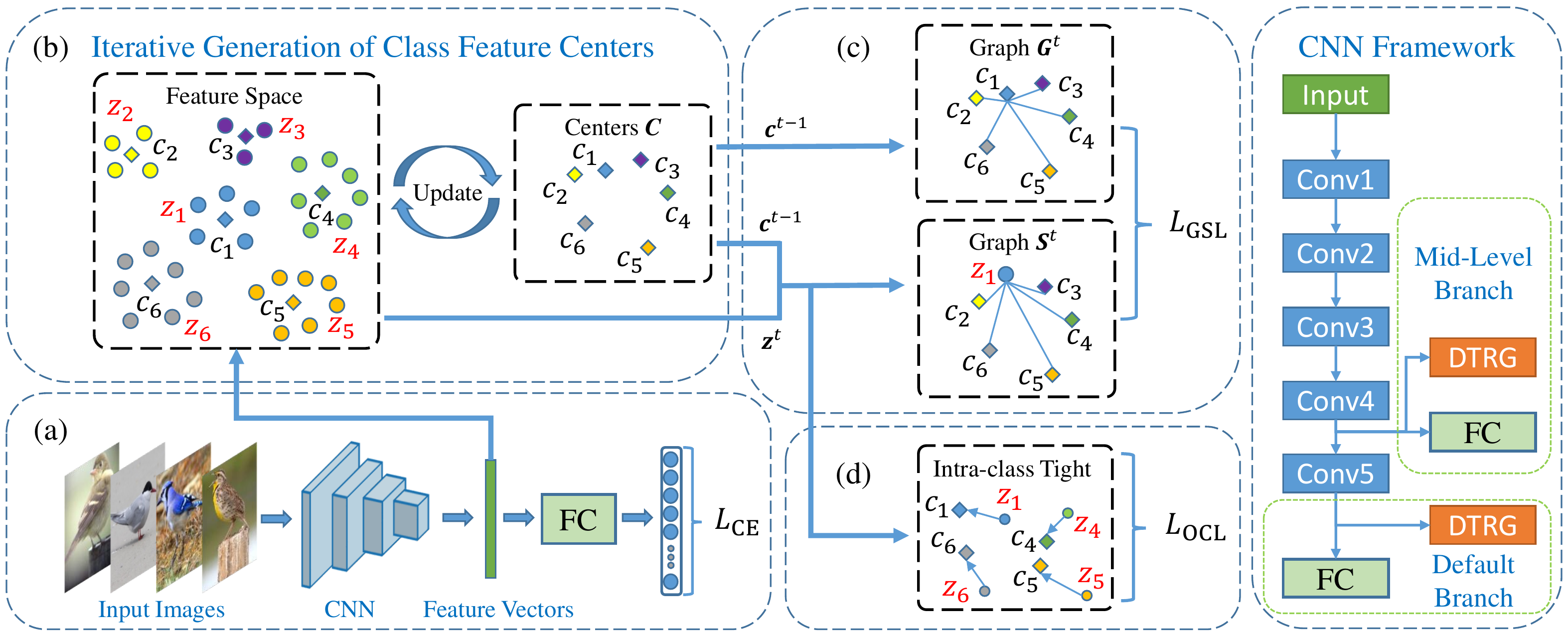}
    \caption{Overview of our proposed Dynamic Target Relation Graphs (DTRG) method. (a) A general backbone for feature extraction and supervised learning, and more details are illustrated in the CNN framework on the right handside. (b) Our module to generate Online Class Feature Centers, where $\bm{z}_i, i=1,2,\ldots,6$ is the sample feature vector and $\bm{c}_i$ denotes the corresponding class feature center. (c) The proposed module to construct inter-class target relation graph, where $\bm{c}^{t-1}$ is the class-level feature from the last epoch (\ie the $(t-1)$-th epoch), $\bm{z}^t$ is the sample feature from current epoch (\ie the $t$-th epoch), the target relation graph $\bm{G}^{t}$, and graph-based representation $\bm{S}^t$. (d) Our Online Center Loss for reducing the intra-class distance. (Best viewed in color)}
    \label{fig:framework}
\end{figure*} 

Given a training set $\{(x_i, y_i)\}_{i=1}^\text{N}$, the problem of fine-grained image classification aims to learn a mapping function $\Phi: \mathcal{X} \rightarrow \mathcal{Y}$ that classifies any test instance into one of $\text{K} = |\mathcal{Y}|$ object categories (\eg bird breeds in fine-grained bird classification \cite{wah2011caltech}), with $x \in \mathcal{X}$ and $y \in \mathcal{Y}$ denoting any input image and its corresponding category label and $\text{N}$ being the number of training instances.
Convolutional neural networks based algorithms \cite{Krizhevsky2012ImageNetCW, He2016DeepRL, Huang2017DenselyCC} have dominated the problem for one decade, which typically can be formulated into a cascade of one feature extractor $\Phi_\text{fea}: \mathcal{X} \rightarrow \mathbb{R}^\text{d}$ and a classifier $\Phi_\text{cls}: \mathbb{R}^{\text{d}}\rightarrow [0,1]^\text{K}$ as
\begin{equation}
\Phi(x) = \Phi_\text{cls}(\bm{z}) \circ \Phi_\text{fea}(x)
\end{equation} 
where $\text{d}$ denotes the dimension of the feature representation output $\bm{z} \in \mathcal{Z}$ of $\Phi_\text{fea}(x)$.
The feature vector can be further mapped to a $\text{K}$-dimensional prediction vector $\bm{p} = \Phi_\text{cls}(\bm{z})=[p_1, p_2, \cdots, p_\text{K}]$ by the classifier.

Dependency in the label space has been verified its effectiveness on improving feature 
discrimination such as label relation graph \cite{Deng2014LargeScaleOC}, label smoothing \cite{zhang2020delving}, target codes \cite{Yang2015DeepRL}.
However, existing methods adopted a static inter-class relation via professional prior knowledge or a soft labeling strategy, which can be less flexible to reveal complex correlation across categories.  
As a result, this paper concerns on discovering latent correlation in the label space and thus proposes a novel dynamic target relation graph constraint on feature encoding $\Phi_\text{fea}(x)$, which correspondingly update the interrelation graph between target classes compounded with feature learning.

The whole pipeline of deep representation learning regularized by dynamic target relation graphs is shown in Fig. \ref{fig:framework}, which can be divided into the backbone of feature encoding (typically the ResNet-50 \cite{He2016DeepRL}, the ResNet-101 \cite{He2016DeepRL}, the DenseNet-161 \cite{Huang2017DenselyCC} in our experiments), generation of dynamic target relation graphs (see Sec. \ref{subsec: graphConstruction}), supervised representation learning using an auxiliary target relation aware constraint (see Sec. \ref{subsec: regularization}).  
More specifically, generation of the dynamic graphs of inter-class relation can have the following two steps: 1) online generation of representation centers (illustrated in Fig. \ref{fig:framework} (b)) and 2) graph construction on the anchored centers (illustrated in Fig. \ref{fig:framework} (c)). 
{{Beyond the popular class supervision using the cross entropy (CE) loss $L_\text{CE}$ as follows:}}
\begin{equation}
\label{eq:celoss}
    L_\text{CE} = -\frac{1}{\text{N}} \sum_{i=1}^\text{N} \sum_{j=1}^\text{K} y_{i,j} \log\frac{\exp(p_{i, j})}{\sum_j \exp(p_{i, j})},
\end{equation}
the aforementioned two modules are additionally supervised to minimize the intra-class variation and model inter-class relation respectively in our scheme, which thus encourage discriminative features.

During testing, an unseen image is fed into a cascade of feature encoder $\Phi_\text{fea}(x)$ and a classifier $\Phi_\text{cls}$ to predict its class probability directly.  

\subsection{Construction of Dynamic Target Relation Graphs}\label{subsec: graphConstruction}

\subsubsection{Generation of class-level Feature Centers}
Before discovering the relation between target classes, the class-level representation (\ie class center in feature space) plays an important role, as it is sensitive to data distribution.
In \cite{wen2016discriminative}, a learnable scheme of adaptive generation of class-wise feature centers is introduced to formulate the generation into parametric estimation. 
Such a method can be effective to minimize intra-class variations, but optimization on those center parameters suffers from the challenge of sparse and imbalanced data distribution in practice.   
The aforementioned observation encourages us to adopt the mean feature vector of samples in each class as the class-level representation, which is robust and easy to obtain during the iteration of parameters updating. 

Technically, we first define the class feature center matrix $\bm{C}^t \in \mathbb{R}^{\text{K}\times \text{d}}$ for $\text{K}$ classes, where the representation of each class is a $\text{d}$-dimensional vector and the superscript $t$ denotes the $t$-th epoch during model training. 
The center matrix $\bm{C}^0$ can be typically initialized for all classes with random Gaussian parameters, which indicates that the centers for all classes are random points in the feature space.
The class-level center matrix $\bm{C}^t$ is iteratively updated by averaging all the features of samples from the same class during every epoch.    
The center matrix $\bm{C}$ can be employed to define a non-parametric online center loss (OCL):
\begin{equation}
\label{eq:centerloss}
    L_\text{OCL} = \frac{1}{\text{N}} \sum_{i=1}^\text{N} \|\bm{z}_i - \bm{C}^{t-1}_{y_i} \|_2^2,
\end{equation}
where $\bm{z}_i$ is the feature vector of any sample and {{$\bm{C}_{y_i}^{t-1}$} is its corresponding class-level representation vector from the center matrix $\bm{C}$ generated in the last epoch (\ie $(t-1)$-th epoch)}. $\|\cdot\|_2$ denotes the Euclidean norm. More details are given in Algorithm \ref{alg:centers}.
\textcolor{black}{
Note that, our generation of class-level representations is similar to prototype learning proposed in \cite{Snell2017PrototypicalNF}.
However, our method is designed in a non-parametric manner, whereas prototype learning usually requires optimization of more learnable parameters and thus can be more sensitive to data distributions.} 

\begin{algorithm}[htp]
    \caption{The pipeline of generating dynamic class centers}
    \label{alg:centers}
    \begin{algorithmic}[1]
    \REQUIRE Dataset $\mathcal{D}_{train} = \{x_i, y_i\}^\text{N}$, feature extractor $\Phi_\text{fea}$, classifier $\Phi_\text{cls}$, training epochs $\text{T}$, class number $\text{K}$, feature dimension $\text{d}$, starting epoch $\text{T}_\text{th}$ to use the total loss (\ref{eq:total loss})\\
    \ENSURE Class feature center matrix $\bm{C}^0 \in \mathbb{R}^{\text{K} \times \text{d}}$ with random variables i.i.d. from a Gaussian distribution
    \FOR{current epoch t = 1 \textbf{to} $\text{T}$}
        \STATE Initialize a sum matrix $\bm{M}\in \mathbb{R}^{\text{K} \times \text{d}}$ and a counting vector $\bm{V}\in \mathbb{R}^{\text{K} \times 1}$ both with zero entries.
        \FOR{iteration = 1 \textbf{to} iterations}
            \STATE Randomly sampling a batch of training data $\mathcal{B} \subset \mathcal{D}_{train}$
            \FOR{$i = 1$ \textbf{to} $|\mathcal{B}|$}
                \STATE Obtain prediction {${\bm{z}_i = \Phi_\text{fea}(x_i), x_i \in \mathcal{B}}$}
                \STATE Update $\bm{M}[y_i]$ and $\bm{V}[y_i]$ with $\bm{M}[y_i] + \bm{z}_i$ and $ \bm{V}[y_i] + 1$ respectively
                \STATE  Obtain prediction $\bm{p}_i = \Phi_\text{cls}(\bm{z}_i)$
            \ENDFOR
            \IF {$t<\text{T}_\text{th}$} 
            \STATE Compute the total loss according to \textbf{Eq.}(\ref{eq:celoss}) 
            \ELSE 
            \STATE Compute the total loss according to \textbf{Eq.}(\ref{eq:total loss}) 
            \ENDIF
            \STATE Backward to update the parameters of the model  
        \ENDFOR
        \FOR{j=1 \textbf{to} $\text{K}$} 
            \STATE Update ${\bm{C}^t}[j] \leftarrow \bm{M}[j] / \bm{V}[j]$
        \ENDFOR
    \ENDFOR
    \textcolor{black}{\RETURN The learned parameters of $\Phi_\text{fea}$ and $\Phi_\text{cls}$}
\end{algorithmic}
\end{algorithm}
    
\subsubsection{Construction of Inter-Class Relation Graphs in Target Space}

Since different categories are always of varying degrees of relation correlation, especially in our fine-grained visual task, there always exist some extremely similar categories.
Given the feature centers for all categories at each epoch, we can construct a target relation graph by measuring relative distance of class-level representations, which is considered as a self-supervised supervision for regularization of deep representation learning with underlying manifolds in the feature space.
Such a setting constrains optimization of feature encoding towards awareness of sample distribution via simplifying with relative positioning of class-level representations, which can thus be robustness against outliers.

To measure target relation with relative distance of class-level feature centers, we firstly consider the Cosine Similarity function as follows:
\begin{equation}\label{eq:sim}
        \cos(\bm{z}_i, \bm{z}_j) = \frac{\bm{z}_i \cdot \bm{z}_j}{\|\bm{z}_i\|_2 \|\bm{z}_j\|_2}.
\end{equation}
To keep the monotonicity and non-negativity of similarity metric, we further define our similarity function as follows:    
    \begin{equation}
    \label{eq:simlarity}
        s(\bm{z}_i, \bm{z}_j) = e^{\hat{\bm{z}}_i \cdot \hat{\bm{z}}_j/\tau},
    \end{equation}
where $\hat{\bm{z}}_i = \frac{\bm{z}_i}{\|\bm{z}_i\|_2}$ is the normalized feature of $\bm{z}_i$ and $\tau$ is a temperature parameter for controlling the variance.
It is noted that other distance metrics such as {{the squared Euclidean distance and the Gaussian kernel function}} can be employed, but the cosine similarity can perform better to discover a similar distribution of feature dimensions.
As a result, we employ \textbf{Eq.} (\ref{eq:simlarity}) in our scheme. 

In mathematics, given a feature center matrix $\bm{C}^{t-1}$ for updating features in the $t$-th epoch, the $j$-th row $\bm{C}_{j} \in \mathbb{R}^\text{d}$ of $\bm{C}$ being the mean feature center within the corresponding category, where $j=1,2,\ldots, \text{K}$ and $t = 1,2,\ldots, \text{T}$, the target relation graph matrix $\bm{G}^{t} \in \mathbb{R}^{\text{K} \times \text{K}}$ can thus be constructed as the following:
    \begin{equation}
    \label{eq: targetGraph}
        \bm{G}_{kl}^{t} = s(\bm{C}_k^{t-1}, \bm{C}_l^{t-1}) = e^{\hat{\bm{C}}_k^{t-1} \cdot \hat{\bm{C}}_l^{t-1}/\tau},
    \end{equation}
where $\bm{G}_{kl}^{t}$ denotes the similarity connected between any two iteration-varying centers $\bm{C}_k^{t-1}$ and $\bm{C}_l^{t-1}$, as we obtain the feature centers using features at the $(t-1)$-th epoch; and $\bm{G}^{t}$ is a dynamic target relation graph.

Having the dynamic self-generated regularization objective (\ie $\bm{G}$) at each epoch,
the samples $x$ are first fed into the deep feature encoder $\Phi_\text{fea}(x)$ to generate their features $\bm{z}$, which are then transformed into graph representations by computing relative distance between these samples' features and feature centers of all categories.  
Specifically, the feature $\bm{z}_i^t=\Phi_\text{fea}(x_i^t)$ for sample $x_i$ at the $t$-the epoch, we can obtain the graph representation of the sample  $\bm{S}^t(\bm{z}_i)$ between the sample feature $\bm{z}_i$ and any centers $\bm{C}_j^{t-1}$ generated from last epoch (the $(t-1)$-th epoch). 
    \begin{equation}
    \label{eq: sampleGraph}
        \bm{S}_{j}^t(\bm{z}_i^t) = s(\bm{z}_i^t, \bm{C}_j^{t-1})  = e^{\hat{\bm{z}}_i^t \cdot \hat{\bm{C}}_j^{t-1}/\tau},
    \end{equation}
where $i \in [1, 2, \ldots, \text{N}]$, $j \in [1, 2, \ldots, \text{K}]$ and epoch $t \in [1, 2, \ldots, \text{T}]$. 
In this way, the graph representation $\bm{S}_i^t = [\bm{S}_{1}^t(\bm{z}_i^t), \bm{S}_{2}^t(\bm{z}_i^t),\ldots, \bm{S}_{j}^t(\bm{z}_i^t), \ldots, \bm{S}_{\text{K}}^t(\bm{z}_i^t)] \in \mathbb{R}^{\text{K}}$  can be generated from the sample's feature $\bm{z}_i^t$ and the feature center matrix $\bm{C}^{t-1}$.

\subsection{Feature Regularization with Target Relation Graphs}\label{subsec: regularization}

In the previous section, the target relation graph $\bm{G}^t$ and the graph-based representation $\bm{S}_i^t$ from $\bm{z}_i$ of sample $x_i$ at the $t$-th epoch are generated respectively.  
We consider maximizing these similarities of class-relation graph by minimizing the distance between target relation graph $\bm{G}_{y_i}^{t}$ and inter-class graph representation $\bm{S}_i^t$ of $x_i$, where $\bm{G}_{y_i}^{t} \in \mathbb{R}^{\text{K}}$ is the vector to model target correlation anchored on label $y_i$ of sample $x_i$.

To measure distance between graphs, we first choose the popular squared Euclidean distance based on the theory analysis from \cite{dubey2018pairwise}.
We can formulate the similarity constraint loss function of inter-class relation graph -- the graph similarity loss (GSL) as follows: 
    \begin{equation}
    \label{eq:euclidean}
        L_\text{GSL}= \frac{1}{\text{N}} \sum_{i=1}^\text{N} \|\bm{S}_i^t - \bm{G}_{y_i}^{t}\|_2^2.
    \end{equation}

Alternatively, the Kullback-Leibler (KL) Divergence is adopted as similarity metric as:
    \begin{equation}
        \mathcal{D}_\text{KL}(P||Q) = \sum_{x \in \mathcal{X}} P(x) \ln \frac{P(x)}{Q(x)}, 
    \end{equation}
where $P$ and $Q$ are probability distributions from the same space $\mathcal{X}$. 
Before using the KL Divergence to encode relation graphs into similarity distance, both relation graph vectors $\bm{S}_i^t$ and $\bm{G}_{y_i}^t$ are normalized by their norms being 1. 
As a result, the normalized relation graphs are depicted as:
    \begin{equation}
        \hat{\bm{S}}_{ij}^t = \frac{\bm{S}_{ij}^t}{\sum_j \bm{S}_{ij}^t}  = \frac{e^{\hat{\bm{z}}_i^t \cdot \hat{\bm{c}}_j^{t-1}/\tau}}{\sum_j e^{\hat{\bm{z}}_i^t \cdot \hat{\bm{c}}_j^{t-1}/\tau}},
    \end{equation}
        
    \begin{equation}
        \hat{\bm{G}}_{ij}^t = \frac{\bm{G}_{ij}^t}{\sum_j \bm{G}_{ij}^t}  = \frac{e^{\hat{\bm{c}}_i^t \cdot \hat{\bm{c}}_j^{t-1}/\tau}}{\sum_j e^{\hat{\bm{c}}_i^t \cdot \hat{\bm{c}}_j^{t-1}/\tau}}.
    \end{equation}

In this way, the KL divergence between $\bm{G}_{y_i}^{t}$ and sample graph $\bm{S}_i^t$ can be formulated as the following:
    \begin{equation}
        \mathcal{D}_\text{KL}(\hat{\bm{G}}_{y_i}^{t}||\hat{\bm{S}}_{i}^t) = \sum_{j=1}^\text{K} \hat{\bm{G}}_{y_i j}^{t} \ln \frac{\hat{\bm{G}}_{y_i j}^{t}}{\hat{\bm{S}}_{ij}^t},
    \end{equation}
\textcolor{black}{where $\hat{\bm{S}}_i^t$ denotes the $i$-th normalized sample graph and $\hat{\bm{G}}_{y_i}^{t}$ is its corresponding target graph.  }   
The graph similarity loss function based on the KL Divergence can thus be written as:
    \begin{equation}
    \label{eq:kl_div}
        L_\text{GSL}= \frac{1}{\text{N}} \sum_{i=1}^\text{N}  \mathcal{D}_\text{KL}(\hat{\bm{G}}_{y_i}^{t}||\hat{\bm{S}}_{i}^t).   
    \end{equation}
{{Both of the Square Euclidean distance and the KL-divergence can be used to constrain the similarity of inter-class relation graph, which are compared in our experiments (see Sec. \ref{subsec: ablation}) for evaluation on the better choice of the graph similarity loss.}}

Beyond the conventional one-hot classification supervision signal, with the loss $L_\text{GSL}$ as an extra self-generated objective, the deep representation learning can be regularized to incorporate target relation.
\textcolor{black}{
It is worth mentioning here that, our graph similarity constraint can also be regarded as a kind of self-distillation method via self-supervised target relation regularization, which is different from traditional graph distillation \cite{Liu2019KnowledgeDV, Lee2019GraphbasedKD, Chen2021LearningSN} that always requires a larger model for teacher graph learning to guide a student network.
}

\subsection{Inter-Class Relation Augmentation}
\label{subsec: relation}

To further enrich the inter-class relation, we introduce an inter-class relation augmentation.
Our inter-class relation augmentation is inspired by recent success of the classical data augmentation \textbf{Mixup} \cite{zhang2018mixup}. 
Specifically, the Mixup arguments more training images by linearly interpolating two random input images $x_a, x_b$ from different classes and their corresponding labels $y_a, y_b$, which can be described as:
    \begin{equation}
    \label{eq:mixup1}
    \begin{aligned}
        &\Tilde{x}_i = (1 - \lambda) x_a + \lambda x _b, \\
        &\Tilde{y}_i = (1 - \lambda) y_a + \lambda y_b,
    \end{aligned}
    \end{equation}
where $\lambda \sim Beta(\alpha, \alpha)$, for $\alpha \in (0, \infty)$. 
\textcolor{black}{Note that, a number of mixup-style data augmentation methods such as the CutMix \cite{Yun2019CutMixRS} and the SnapMix\cite{huang2021snapmix} can also be depicted by \textbf{Eq.} (\ref{eq:mixup1}), while they just differ from the way to generate $\lambda$. 
As a result, these mixup augmentation strategies are adopted in our inter-class relation augmentation, whose performance differences are compared in our experiments (See Table \ref{table:mixup-snapmix}).
}
   
We follow the same data augmentation processing as the Mixup \cite{zhang2018mixup} to generate training samples $\{\Tilde{x}_i, \Tilde{y}_i\}^{\text{N}'}$, and then we leverage soft labels of these mixed images for creating more diverse target relation graph. 
Firstly, we obtained the feature $\Tilde{\bm{z}}_i$ of augmented image $\Tilde{x}_i$.
    \begin{equation}
    \label{eq:mixup2}
        \Tilde{\bm{z}}_i = \Phi_\text{fea}(\Tilde{x}_i).
    \end{equation}
For features $\Tilde{\bm{z}}$ from mixed images, we use the pipeline shown in Algorithm \ref{alg:framework} to generate a dynamic class center matrix $\bm{C}^t$. Secondly, we can calculate a mixed version of non-parametric online center loss as:
    \begin{equation}
    \label{eq:mixed centerloss}
        L_\text{OCL} = \frac{1}{\text{N}'} \sum_{i=1}^{\text{N}'} \{(1- \lambda) \|\Tilde{\bm{z}}_i - \bm{C}_{y_a}^{t-1} \|_2^2 + \lambda \|\Tilde{\bm{z}}_i - \bm{C}_{y_b}^{t-1} \|_2^2\}.
    \end{equation}    
\textcolor{black}{
Beyond \textbf{Eq.} (\ref{eq:mixed centerloss}), other mixed options for $L_\text{OCL}$ can be employed such as $\frac{1}{\text{N}'} \sum_{i=1}^{\text{N}'}||\tilde{z}_i -(1-\lambda)\bm{C}_{y_a}^{t-1}-\lambda \bm{C}_{y_b}^{t-1}||_2^2$. Here, we select \textbf{Eq.} (\ref{eq:mixed centerloss}) in our scheme in view of its simple form to make representation learning easy.
}

For generating mixed sample inter-class relation graph $\bm{S}_i^t(\Tilde{\bm{z}}_i)$ and its corresponding target relation graph $\bm{G}_{{y}_a}^{t}$ and $\bm{G}_{{y}_b}^{t}$, the following equations are utilized:
     \begin{equation}
    \label{eq:mixup3}
        \bm{S}_{j}^t(\Tilde{\bm{z}}_i) = s(\Tilde{\bm{z}}_i, \bm{C}_j^{t-1}),
    \end{equation}
    
    \begin{equation}
    \begin{aligned}
    \label{eq:mixup4}
        \bm{G}_{{{y}_a}j}^{t} &= s(\bm{C}_{{y}_a}^{t-1}, \bm{C}_j^{t-		1}), \\
        \bm{G}_{{{y}_b}j}^{t} &= s(\bm{C}_{{y}_b}^{t-1}, \bm{C}_j^{t-1}),
    \end{aligned}
    \end{equation}
where $\bm{C}_{*}^{t-1}= {\bm{C}^{t-1}}[*]$, $j \in [1, 2, ..., \text{K}]$, and $*$ represents the class index.
    
{{
Similarly, we can construct a mixed graph similarity loss based on the former relation graphs in \textbf{Eq.} (\ref{eq:mixup3}) and \textbf{Eq.} (\ref{eq:mixup4}).
}}
Therefore, we use the Square Euclidean distance to formulate the following mixed similarity loss function as:
    \begin{equation}
    \label{eq:mixup5}
        L_\text{GSL} = \frac{1}{\text{N}'} \sum_{i=1}^{\text{N}'} \{(1- \lambda) \|\bm{S}_{i}^t - \bm{G}_{{y}_a}^{t}\|_2^2 + \lambda \|\bm{S}_{i}^t - \bm{G}_{{y}_b}^{t}\|_2^2\}.
    \end{equation}
Note that, we can also construct the KL-divergence based mixed similarity loss function, but superior performance of the Square Euclidean distance to than KL-divergence is achieved in our experiments (see Fig. \ref{fig:parameter} for details). 
Owing to such an adaptation of mixup style data augmentation into the proposed dynamic target relation graph scheme, a diverse data to model complex target relation can further boost performance of visual classification, which is verified in our experiments.
    
\subsection{Model Training}

The total loss for our method can be formulated as:
    \begin{equation}
    \label{eq:total loss}
        L = L_\text{CE} + \beta L_\text{OCL} + \eta L_\text{GSL},
    \end{equation}
where $\beta$ and $\eta$ are trade-off hyper-parameters.
As all terms of the object function in \textbf{Eq.} (\ref{eq:total loss}) are differentiable regardless of selection of two distance metric options for the $L_\text{GSL}$ and whether to use inter-class correlation augmentation.
Consequently, the model can be trained in an end-to-end learning manner with stochastic gradient descent algorithm \cite{Bottou2010LargeScaleML}. 
To learn a relatively stable center matrix $\bm{C}$, we can use a warm-up strategy by training the model only with $L_\text{CE}$ loss function for the first $\text{T}$-{th} epochs, which can make sure that the center matrix to avoid making excessive deviation.
The whole training processing with inter-class relation augmentation is summarized in Algorithm \ref{alg:framework}.

\begin{algorithm}[htb]
    \caption{Model training  processing with Inter-Class Relation Augmentation}
    \label{alg:framework}
    \begin{algorithmic}[1]
    \REQUIRE Dataset $\mathcal{D}_{train} = \{(x_i, y_i)\}$, the feature extractor $\Phi_\text{fea}$, the classifier $\Phi_\text{cls}$, training epochs $\text{T}$, class number $\text{K}$, feature dimension $\text{d}$, start epoch $\text{T}_{th}$, hyper-parameter $\alpha$ for mixup style data augmentation\\
    \ENSURE Class feature center matrix $\bm{C}^0 \in \mathbb{R}^{\text{K} \times \text{d}}$ with random Gaussian parameters 
    \FOR{current epoch t=1 \textbf{to} $\text{T}$}
        \STATE Initialize the middle matrix $\bm{M}\in \mathbb{R}^{\text{K} \times \text{d}}$ and the count matrix $\bm{V}\in \mathbb{R}^{\text{K} \times 1}$ both with zeros.
        \FOR{iteration = 1 \textbf{to} iterations}
            \STATE Random sample a batch training data $\mathcal{B} \subset \mathcal{D}_{train}$
            \FOR{$i = 1$ \textbf{to} $|\mathcal{B}|$}
                \STATE Generate mixed sample $\{\Tilde{x}_i,\Tilde{y}_i\}$ by \textbf{Eq.}(\ref{eq:mixup1})
                \STATE Obtain the predicted feature $\Tilde{\bm{z}}_i$ by \textbf{Eq.}(\ref{eq:mixup2})
                \STATE Update $\bm{M}[y_a] \leftarrow \bm{M}[y_a] + (1 - \lambda) \Tilde{\bm{z}}_i$ and $\bm{M}[y_b] \leftarrow \bm{M}[y_b] + \lambda \Tilde{\bm{z}}_i$
                \STATE Update $\bm{V}[y_a] \leftarrow \bm{V}[y_a] + (1 - \lambda)$ and $\bm{V}[y_b] \leftarrow \bm{V}[y_b] + \lambda$ 
                \STATE Compute the center loss $L_\text{OCL}$ by \textbf{Eq.}(\ref{eq:mixed centerloss})
                \STATE Compute the similarity loss $L_\text{GSL}$ by \textbf{Eq.}(\ref{eq:mixup5})
                \STATE Obtain the prediction vector $\Tilde{\bm{p}}_i = \Phi_\text{cls}(\Tilde{\bm{z}}_i)$
            \ENDFOR
            \IF {$t<\text{T}_{th}$}
                \STATE Compute total loss according by \textbf{Eq.}(\ref{eq:celoss}) 
            \ELSE
                \STATE Compute total loss according by \textbf{Eq.}(\ref{eq:total loss}) 
            \ENDIF
            \STATE Backward to update the parameters of model  
        \ENDFOR
        \FOR{j=1 \textbf{to} $\text{K}$} 
            \STATE Update ${\bm{C}^t}[j] \leftarrow \bm{M}[j] / \bm{V}[j]$
        \ENDFOR
    \ENDFOR
    \textcolor{black}{\RETURN The learned parameters of $\Phi_\text{fea}$ and $\Phi_\text{cls}$}
\end{algorithmic}
\end{algorithm}

\section{Experiments}
\label{sec: exps}

{In this section, to evaluate the effectiveness and generalization of our proposed method, we conduct experiments on diverse public benchmarks of multiple visual classification tasks.
We firstly investigate details of datasets and experimental settings. 
Secondly, the proposed method and the state-of-the-art methods are evaluated and compared on popular benchmarks.
Thirdly, we verify superior robustness of our method for sparse and imbalanced data. 
Moreover, ablation studies for the proposed components are evaluated and compared, to verify their effectiveness.
Finally, visualizations based on t-SNE \cite{Maaten2008VisualizingDU} and Grad-CAM \cite{Selvaraju2019GradCAMVE} are presented for better understanding our motivation.
}

\subsection{Datasets and Experimental Settings}
\label{subsec: setting}
\begin{figure}
    \centering
    \includegraphics[width=\linewidth]{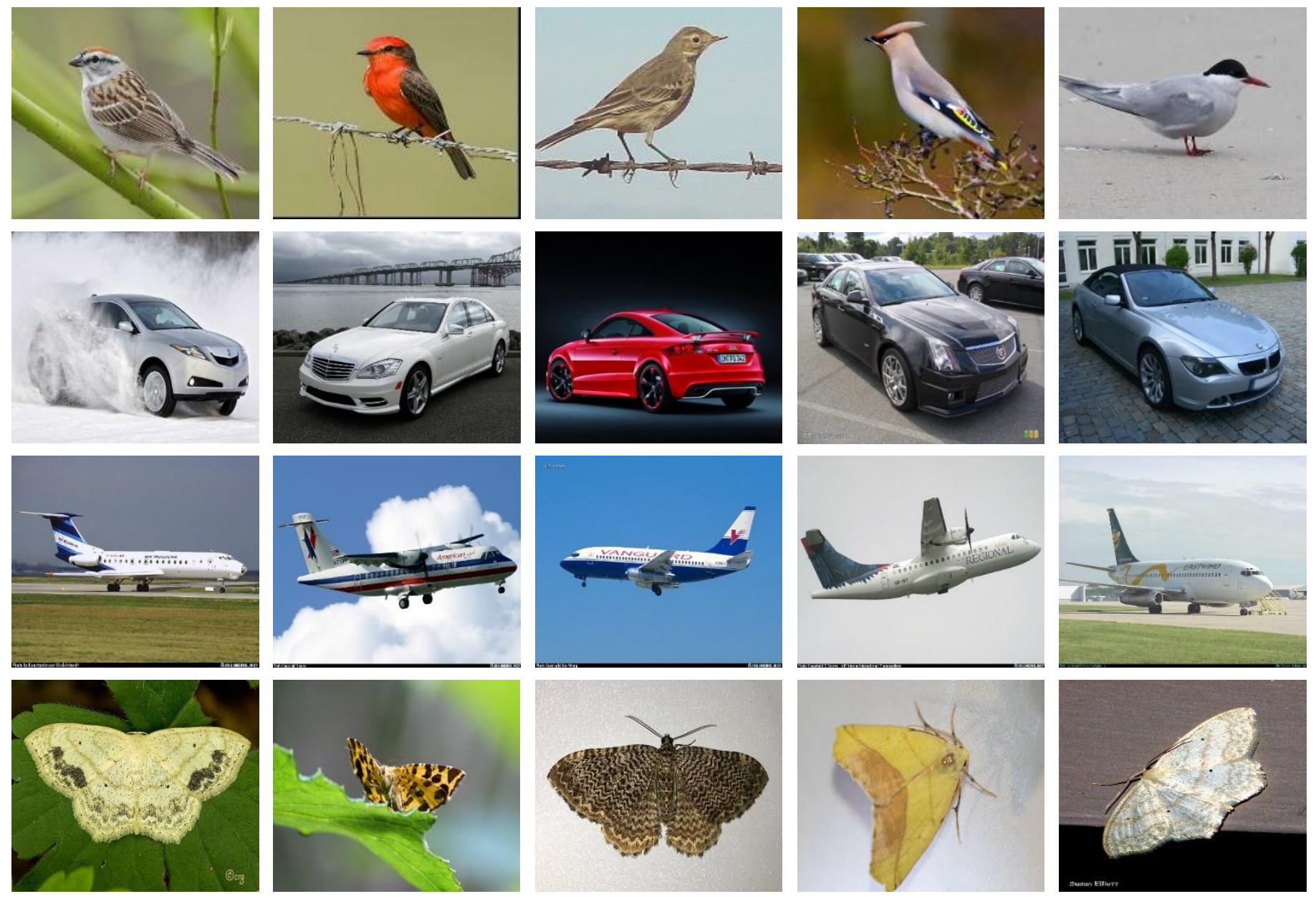}
    \caption{Examples from four fine-grained datasets. Images in the top row are from the CUB-200-2011 \cite{wah2011caltech}, those in the second row are from the Stanford Cars \cite{krause20133d}, those in the third row are from the FGVC Aircraft \cite{maji2013fine}, and those in the bottom row are from the iNaturalist 2018 \cite{Horn2018TheIS}.
    }
    \label{fig:sample images}
\end{figure}

{In the problem of fine-grained image classification,} we evaluate our method on the following widely used datasets: the CUB-200-2011 (CUB) \cite{wah2011caltech}, the Stanford Cars (CAR) \cite{krause20133d}, the FGVC Aircraft (AIR) \cite{maji2013fine}, and the large-scale long-tailed iNaturalist 2018 \cite{Horn2018TheIS}. 
In our experiments, we follow the data splits provided in  \cite{huang2021snapmix}, whose details are provided in Table \ref{table:datasets}, while the example images are visualized in Fig. \ref{fig:sample images}.
\textcolor{black}{
More experiments are conducted on the problem of generic image and texture classification, on two generic image classification benchmarks -- the CIFAR \cite{krizhevsky2009learning} and the ImageNet \cite{deng2009imagenet} and also two texture image datasets -- the DTD \cite{Cimpoi2014DescribingTI} and the MIT Scene (MIT) \cite{Quattoni2009RecognizingIS}.
The dataset statistics are also summarized in Table \ref{table:datasets}, and training/testing splits for the DTD and the MIT datasets from their official websites are adopted in our experiments.
Only category labels of training data are adopted as supervision signals without using any additional prior information.}

\begin{table}[t]
\centering
\caption{\textcolor{black}{The statistics of datasets.}}
\label{table:datasets}
\setlength{\tabcolsep}{2mm}
\begin{tabular}{@{}lccc@{}}
\toprule
Dataset       & \# classes   & \# training data & \# testing data \\ \midrule
CUB \cite{wah2011caltech}  & 200      & 5,994      & 5,794  \\
CAR \cite{krause20133d}  & 196       & 8,144      & 8,041  \\
AIR \cite{maji2013fine}   & 100      & 6,667      & 3,333  \\ 
iNaturalist 2018 \cite{Horn2018TheIS} & 8142  & 437,513  & 24,426   \\ \midrule\midrule
CIFAR10/100 \cite{krizhevsky2009learning}         & 10/100    & 50,000   & 10,000   \\
ImageNet  \cite{deng2009imagenet}             & 1000  & 1,281,167 & 50,000   \\\midrule\midrule
DTD \cite{Cimpoi2014DescribingTI}             & 47    & 1,880    & 1,880    \\
MIT  \cite{Quattoni2009RecognizingIS}         & 67    & 5,360    & 1,340    \\\bottomrule
\end{tabular}
\end{table}

\vspace{0.1cm}\noindent\textbf{Data Pre-Processing --} 
The raw images from the CUB are firstly resized to $512 \times M$ without warping, where $M$ is the length of a longer side, while images from the CAR, the AIR, the DTD and the MIT datasets are resized to $512 \times 512$ directly with warping, following the same setting in \cite{huang2021snapmix, Chang2020TheDI}.
These images are then cropped with size $448 \times 448$ for batch training and testing. 
\textcolor{black}{For the ImageNet and the iNaturalist 2018 datasets, the images are cropped to size $224 \times 224$.
During training, we adopt random cropping and random horizontal flipping for data augmentation by following existing works \cite{huang2021snapmix, Chang2020TheDI,Park2021InfluenceBalancedLF}; while we only adopt center cropping during testing as \cite{huang2021snapmix}.
\textcolor{black}{For the CIFAR datasets, we randomly crop the images into size $32 \times 32$ with 4 pixels padding at each side and with random horizontal flipping by following the basic data augmentation operations used in \cite{He2016DeepRL, zhang2020delving}.}}

\begin{table}[t]
\centering
\caption{Comparison with state-of-the-art methods on classification accuracy ($\%$) with the CUB, the CAR and the AIR. 
The reported results of our method are obtained by combining our proposed \textbf{dynamic target relation graph (DTRG)} with \textbf{baseline+mid} and \textbf{mixup-based data augmentation}. The best results for each dataset are marked in bold.}
\label{table:sota}
\setlength{\tabcolsep}{1mm}
\begin{tabular}{@{}lccccc@{}}
\toprule
Method        & Base Model   & Year  & {CUB (\%)} & {CAR (\%)} & {AIR (\%)}       \\ \midrule
Bilinear-CNN\cite{lin2015bilinear}    & VGG16       & 2015 & 84.1          & 91.3          & 84.1          \\
MoNet\cite{Gou2018MoNetME}	          & VGG16	    & 2018	& 86.4	        & 91.8	        & 89.3          \\
HBP\cite{yu2018hierarchical}          & VGG16       & 2018 & 87.1          & 93.7          & 90.3          \\ \midrule

MGD \cite{Wang2015MultipleGD}         & VGG19       & 2015 & 81.7          & -             & 82.5          \\
RA-CNN\cite{fu2017look}               & VGG19       & 2017 & 85.3          & 92.5          & -             \\
MA-CNN\cite{zheng2017learning}        & VGG19       & 2017 & 86.5          & 92.8          & 89.9          \\ \midrule

iSQRT-COV\cite{Li2018TowardsFT}	      & ResNet50	& 2018	& 88.1	               & 92.8	& 90.0          \\
DFL\cite{Wang2018LearningAD}          & ResNet50    & 2018 & 87.4          & 93.1          & 91.7          \\
DCL\cite{chen2019destruction}         & ResNet50    & 2019 & 87.8          & 94.5          & 93.0            \\
LIO\cite{zhou2020look}                & ResNet50    & 2019 & 88.0            & 94.5          & 92.7          \\
S3N\cite{ding2019selective}           & ResNet50    & 2019 & 88.5          & 94.7          & 92.8          \\
MGE-CNN\cite{Zhang2019LearningAM}    & ResNet50    & 2019 & 88.5          & 93.9          & -             \\
MC-Loss\cite{Chang2020TheDI}          & ResNet50    & 2020 & 87.3          & 93.7          & 92.6          \\

PMG\cite{du2020fine}                  & ResNet50    & 2020 & \textbf{89.6} & 95.1          & 93.4          \\

GaRD\cite{Zhao2021GraphbasedHR}	& ResNet50	& 2021	& \textbf{89.6}	       & 95.1	& \textbf{94.3}          \\
SVD-Trunc\cite{Song2021WhyAM}	& ResNet50	& 2021	& 87.3	                & 93.0	& -             \\
SPS\cite{Huang2021StochasticPS}	& ResNet50	& 2021	& 88.7	                & 94.9	& 92.7          \\

DTRG (Ours)                            & ResNet50    & -    & 88.8         & \textbf{95.2}         & {94.1}         \\ \midrule

iSQRT-COV\cite{Li2018TowardsFT}	        & ResNet101	& 2018	& 88.7	         & 93.3     	& 91.4          \\
DBTNet\cite{Zheng2019LearningDB}	   & ResNet101	& 2019	& 88.1          & 94.5          & 91.6 \\
CIN\cite{Gao2020ChannelIN}            & ResNet101   & 2020 & 88.1          & 94.5          & 92.8          \\
SnapMix\cite{huang2021snapmix}        & ResNet101   & 2021 & 89.3          & 94.8          & 94.1         \\
DTRG (Ours)                            & ResNet101   & -   & \textbf{89.7} & \textbf{95.4} & \textbf{94.2} \\ \midrule

PC\cite{dubey2018pairwise}            & DenseNet161  & 2018 & 86.9         & 92.9          & 89.2         \\
FDL\cite{Liu2020FiltrationAD}	       & DenseNet161 & 2020		& 89.1	& 94.0	& 91.3\\
API-Net\cite{zhuang2020learning}      & DenseNet161  & 2020 & {90.0}       & {95.3}        & 93.9         \\ 
DTRG (Ours)                            & DenseNet161 & -    & \textbf{90.1}& \textbf{95.7}& \textbf{94.8}\\ \bottomrule
\end{tabular}
\end{table}

\vspace{0.1cm}\noindent\textbf{Backbone and Baseline --} {For fine-grained image classification on the CUB, the CAR and the AIR,} to better evaluate the proposed DTRG method, three popular networks are employed as backbone, \ie the ResNet50 \cite{He2016DeepRL}, the ResNet101 \cite{He2016DeepRL} and the DenseNet161 \cite{Huang2017DenselyCC}, which are all pre-trained on the ImageNet \cite{deng2009imagenet}. 
Beyond the backbone networks with only one simple linear classifier as a vanilla model, inspired by several recent works \cite{huang2021snapmix, Zhang2019LearningAM} to boost the performance of recognition, a double-branch strategy is adopted using the CE loss to supervise features from different network depth, as our advanced baseline, which is termed as \textbf{backbone+mid}. 
In our experiments, we follow the implementation from the Sanpmix \cite{huang2021snapmix} by adding an extra branch with another linear classifier on features of the mid-level layer of backbone networks.
Specifically, such a mid-level branch includes a ConvBlock with one $Conv\ 1\times1$ layer and one $ReLU$ module, a $Max\ Pooling$ layer and a $Linear$ Classifier with one layer. 
For the ResNets, this mid-level branch is placed after the $4$-th block of backbone network, while for the DenseNet, that is placed after the penultimate denseblock of backbone network.
By incorporating the new branch in the mid-level level of backbone networks, the double-branch network architecture can capture cross-granularity features. 
Note that, the gradients from the mid-level branch are blocked to the main backbone during training. 
In inference, we use the ensemble of prediction outputs from two classification heads to generate final predictions.
\textcolor{black}{On the iNaturalist 2018, our proposed DTRG is compared with baseline methods using the ResNet50 \cite{He2016DeepRL} as their backbone, as well as several state-of-the-art methods \cite{Cui2019ClassBalancedLB,Cao2019LearningID,Park2021InfluenceBalancedLF}.}
\textcolor{black}{For the CIFAR datasets, the baselines are based on the backbone of the ResNet18/34/50/101 \cite{He2016DeepRL} respectively, while on the ImageNet dataset, our proposed DTRG is compared with the baselines using the ResNet50/101 backbone networks.}
\textcolor{black}{For texture classification on the DTD and the MIT datasets, we compare our method with the baseline based on the pre-trained ResNet50 \cite{He2016DeepRL}. 
}
\textcolor{black}{
Note that, our models for the CIFAR, the ImageNet and the iNaturalist datasets are all trained from scratch.}  

\begin{table}[t]
\centering
\caption{\textcolor{black}{Comparison of classification accuracy (\%) on the iNaturalist 2018 dataset using the ResNet50 backbone.}}
\label{table:inaturalist}
\begin{tabular}{@{}lccc@{}}
\toprule
\multirow{2}{*}{Method} & \multirow{2}{*}{Schedule} & \multicolumn{2}{c}{{iNaturalist 2018}} \\ \cmidrule(l){3-4}               &  & top-1 (\%)                  & top-5 (\%)                  \\ \midrule
Focal\cite{Lin2017FocalLF}                   & SGD                       & 58.03                 & 78.65                 \\
CB-Focal\cite{Cui2019ClassBalancedLB}                & SGD                       & 61.12                 & 81.03                 \\
LDAM\cite{Cao2019LearningID}                    & SGD                       & 64.58                 & 83.52                 \\
IB\cite{Park2021InfluenceBalancedLF}                      & SGD                       & 65.39                 & 84.98                 \\ 
Baseline                & SGD                       & 64.21                 & 84.35                 \\
DTRG (Ours)                      & SGD                       & \textbf{65.54}        & \textbf{85.99}        \\ \midrule\midrule
LDAM\cite{Cao2019LearningID}                    & DRW                       & 68.00                 & 85.18                 \\ 
Baseline                & DRW                       & 67.03                 & 84.75                 \\
DTRG (Ours)                & DRW                       & \textbf{69.47}        & \textbf{87.36}        \\ \bottomrule
\end{tabular}
\end{table}

\vspace{0.1cm}\noindent \textbf{Training Details --} For computational efficiency, in all our {fine-grained and texture recognition} experiments, the models without using mixup-based target relation augmentation are trained for 100 epochs, while only these with mixup-based augmentation are trained for 200 epochs to exploit more diverse target relation for superior performance. 
The hyper-parameter $\alpha$ of \textbf{Eq.} (\ref{eq:mixup1}) is set as 0.1 for the Mixup and 1 for the CutMix and the SnapMix, respectively. 
The parameter $\tau$ in similarity constraint (\textbf{Eq.} (\ref{eq: targetGraph}) and \textbf{Eq.} (\ref{eq: sampleGraph})) is set as 1 for the vanilla backbone and 2 for \textbf{backbone+mid}; while the parameter $\beta$ of \textbf{Eq.} (\ref{eq:total loss}) for controlling the influence of online center loss is set as $1e-3$ by our ablation studies; the parameter $\eta$ of \textbf{Eq.} (\ref{eq:total loss}) is set as 1 for the squared Euclidean Distance by our ablation experiments. 
$T_{th}$ in Algorithms \ref{alg:centers} and \ref{alg:framework} is both set as 2 for better iterating the class feature center matrix during the first epoch, \textcolor{black}{verified by our ablation studies (see Table \ref{table:epoch-ablation})}.
During training, we adopt the stochastic gradient descent (SGD) for network operation with momentum of 0.9 and weight decay of $1e-4$. 
The batch size in all our experiments is set as 16. 
The initial learning rate is set as 0.01 for training from scratch and 0.001 for pre-trained models, which descends by following the multi-step schedule with decay factor as 0.1. 
The milestones for decaying learning rate are set as $[40, 70]$ for 100 epochs and $[80,150,180]$ for 200 epochs.
\textcolor{black}{In the experiments on the CIFAR, we use the SGD optimizer with momentum 0.9 and weight decay $5e-4$ to train all the models for 200 epochs with a batch size of 128. 
The learning rate is initially set as 0.1 and decays at the 100th and 150th epoch by a factor of 0.1, respectively. 
The other hyper-parameters in these experiments are set as same as introduced before.}
\textcolor{black}{In the experiments on the ImageNet and the iNaturalist 2018, we train all the models with a batch size 256, the initial learning rate 0.1, and cosine learning rate schedule for better optimization. 
The training epochs for the ImageNet dataset is set as 100, while 200 for the iNaturalist dataset. 
Due to the large scale size of both two datasets, we set the hyper-parameters $T_{th}$ to be half of the size of total epochs for better generating feature centers. 
The other hyper-parameters $\tau$ is set as 2, $\beta$ is set as $2e-4$, and $\eta$ is set as $2e-3$ for balancing the loss value of our DTRG for the iNaturalist dataset, while $\eta$ is set as $0.1$ on the ImageNet dataset.}

\subsection{Comparative Evaluation}
\vspace{0.1cm}\noindent {\textbf{Evaluation on Fine-grained Classification --}}
Comparative evaluation on the CUB-200-2011, the Stanford Cars, and the FGVC-Aircraft datasets is reported in Table \ref{table:sota}. 
Evidently, our method can consistently achieve superior performance to the state-of-the-art methods on the three  datasets, \ie \textbf{90.1\%}, \textbf{95.7\%} and \textbf{94.8\%} on the CUB, the CAR, and the AIR datasets respectively. 
Moreover, with three popular backbone networks (the ResNet50, ResNet101, and DenseNet161), the proposed DTRG method can always perform better than existing methods, which further verifies its effectiveness.
\textcolor{black}{
Furthermore, to verify the effectiveness of our method on a large-scale fine-grained benchmark, we conduct one more experiment on the iNaturalist 2018 dataset. 
Comparative evaluation with the state-of-the-art methods is reported in Table \ref{table:inaturalist}, where our DTRG method can always outperform the baseline and also be superior to existing works, especially when training with the training schedule DRW introduced in \cite{Cao2019LearningID}.
The results on the iNaturalist again demonstrate our motivation. 
}

\begin{table}[t]
\centering
\caption{\textcolor{black}{Comparative evaluation on classification accuracy (\%) with the Cifar10 and the Cifar100 using different ResNet architectures. The mean and standard deviation are obtained over three trails.}}
\label{table:cifar}
\begin{tabular}{@{}lcccc@{}}
\toprule
\multirow{2}{*}{Backbone} & \multicolumn{2}{c}{Cifar10 (\%)}                & \multicolumn{2}{c}{Cifar100 (\%)}               \\ \cmidrule(l){2-5} 
                          & \multicolumn{1}{c}{Baseline} & DTRG           & \multicolumn{1}{c}{Baseline} & DTRG           \\ \midrule
ResNet18                  & 95.20$\pm$0.08                     & \textbf{95.38$\pm$0.04} & 77.79$\pm$0.09                     & \textbf{78.51$\pm$0.19} \\
ResNet34                  & 95.25$\pm$0.13                     & \textbf{95.61$\pm$0.12} & 77.94$\pm$0.15                     & \textbf{78.83$\pm$0.39} \\
ResNet50                  & 95.24$\pm$0.07                     & \textbf{95.35$\pm$0.10} & 78.49$\pm$0.11                     & \textbf{78.92$\pm$0.16} \\
ResNet101                 & 95.40$\pm$0.06                     & \textbf{95.45$\pm$0.09} & 79.75$\pm$0.13                     & \textbf{80.11$\pm$0.09} \\ \bottomrule
\end{tabular}
\end{table}

\begin{table}[t]
\centering
\caption{\textcolor{black}{Comparative evaluation on classification accuracy (\%) with the ImageNet using the ResNet50 or the ResNet101.}}
\label{table:imagenet}
\begin{tabular}{@{}lccc@{}}
\toprule
\multirow{2}{*}{Backbone}  & \multirow{2}{*}{Method} & \multicolumn{2}{c}{ImageNet} \\ \cmidrule(l){3-4} 
                           &                         & top-1 (\%)             & top-5 (\%)             \\ \midrule
\multirow{2}{*}{ResNet50}  & Baseline                & 76.63             & 93.16             \\
                           & DTRG                    & \textbf{77.11}    & \textbf{93.29}    \\ \midrule
\multirow{2}{*}{ResNet101} & Baseline                & 78.41             & 	94.22             \\
                           & DTRG                    & \textbf{78.87}    & \textbf{94.31}     \\  \bottomrule
\end{tabular}
\end{table}

\vspace{0.1cm}\noindent \textcolor{black}{\textbf{Evaluation on Ordinary Image Classification --} As our method is not limited to fine-grained classification, we evaluate the proposed DTRG on the CIFAR10/100 and the ImageNet of general image classification problem with different backbone architectures, respectively. 
Comparative evaluation of different ResNet variants with or without our DTRG method on the CIFAR10/100 is reported in Table \ref{table:cifar}, where our DTRG can consistently beat the baseline methods. 
Similar results on the large-scale ImageNet in Table \ref{table:imagenet}
can be observed with the ResNet50/101 as the backbone.
The results on all datasets can confirm that our DTRG is generic and can be inserted in existing deep networks for different visual classification tasks.}

\vspace{0.1cm}\noindent \textcolor{black}{\textbf{Evaluation on Texture Classification --} 
Comparative evaluation on the DTD and the MIT datasets is reported in Table \ref{table:dtd-mit}. 
Superior performance of our DTRG with the ResNet50 backbone can be achieved to the baseline and the state-of-the-art methods, \ie the TKPF\cite{yu2022efficient} and the SPS \cite{Huang2021StochasticPS}. 
On the DTD, we report the mean accuracy across the 10 splits provided in \cite{Cimpoi2014DescribingTI}, and our method can outperform the baseline with a significant margin, \ie by about \textbf{3\%}. 
The results on the two texture datasets demonstrate that our methods can improve the representation learning on diverse visual classification tasks, specifically fine-grained scene recognition and texture classification.} 

\begin{table}[t]
\centering
\caption{\textcolor{black}{Comparison of classification accuracy (\%) with state-of-the-art methods on the DTD and the MIT Scene datasets.}}
\label{table:dtd-mit}
\begin{tabular}{@{}lccc@{}}
\toprule
Method   & Base Model & {DTD} (\%)       & {MIT} (\%)   \\ \midrule
CBP-RM\cite{Gao2016CompactBP}   & VGG16      & 64.5                  & 76.2 \\
RUN\cite{Yu2020TowardFA}        & VGG16      & 68.4                 & 80.8 \\
TKPF\cite{yu2022efficient}      & VGG16      & 68.2                 & 80.5 \\ \midrule
ReDro\cite{Rahman2020ReDroEL}   & ResNet50   & -                    & 84.0   \\
SPS\cite{Huang2021StochasticPS} & ResNet50   & -                     & 83.1 \\
TKPF\cite{yu2022efficient}      & ResNet50   & 71.4                 & 84.1 \\ 
Baseline                        & ResNet50   & 69.2$\pm$0.5             & 83.4 \\
DTRG (Ours)                     & ResNet50   & \textbf{72.1$\pm$0.7} & \textbf{84.2} \\ \bottomrule
\end{tabular}
\end{table}

\begin{table}[t]
\centering
\caption{\textcolor{black}{Comparison of classification accuracy (\%) with the Label Smooth, the OLS and the Center Loss methods on the CUB, the CAR and the AIR datasets using the ResNet50 backbone.}}
\label{table:center-loss}
\begin{tabular}{@{}lccc@{}}
\toprule
Method       & CUB (\%)      & CAR (\%)      & AIR (\%)      \\ \midrule
Label Smooth \cite{Szegedy2016RethinkingTI} & 86.0          & 93.4          & \textbf{92.0}              \\
OLS \cite{zhang2020delving}          & 86.3          & 93.3          & 91.2              \\ 
Center Loss \cite{wen2016discriminative}  & 86.4          &93.5      & 91.4              \\
DTRG (Ours)  & \textbf{87.3} & \textbf{94.3} & \textbf{92.0} \\ \bottomrule
\end{tabular}
\end{table}

\vspace{0.1cm}\noindent \textcolor{black}{
\textbf{Comparison with Label Smooth and Center Loss --}
As the label smooth and the center loss algorithms are our direct competitors, for a fair comparison of our DTRG method with the vanilla label smoothing \cite{Szegedy2016RethinkingTI}, the online label smooth (OLS) \cite{zhang2020delving} and the Center Loss \cite{wen2016discriminative}, we conduct an experiment on the CUB-200-2011, the Stanford Cars, and the FGVC-Aircraft datasets with the ResNet50 backbone under the same settings. 
Results are illustrated in Table \ref{table:center-loss}. 
It is observed that our DTRG method can achieve the best performance among all comparative methods. 
These results can confirm again the superiority of our DTRG method to these competing methods.
Moreover, the better performance of our DTRG method can be gained because of non-parametric center generation, which can be more stable than Center Loss \cite{wen2016discriminative} trained with additional learnable parameters.
}

\begin{figure}[t]
    \centering
    \includegraphics[width=0.98\linewidth, trim=20 5 20 45,clip]{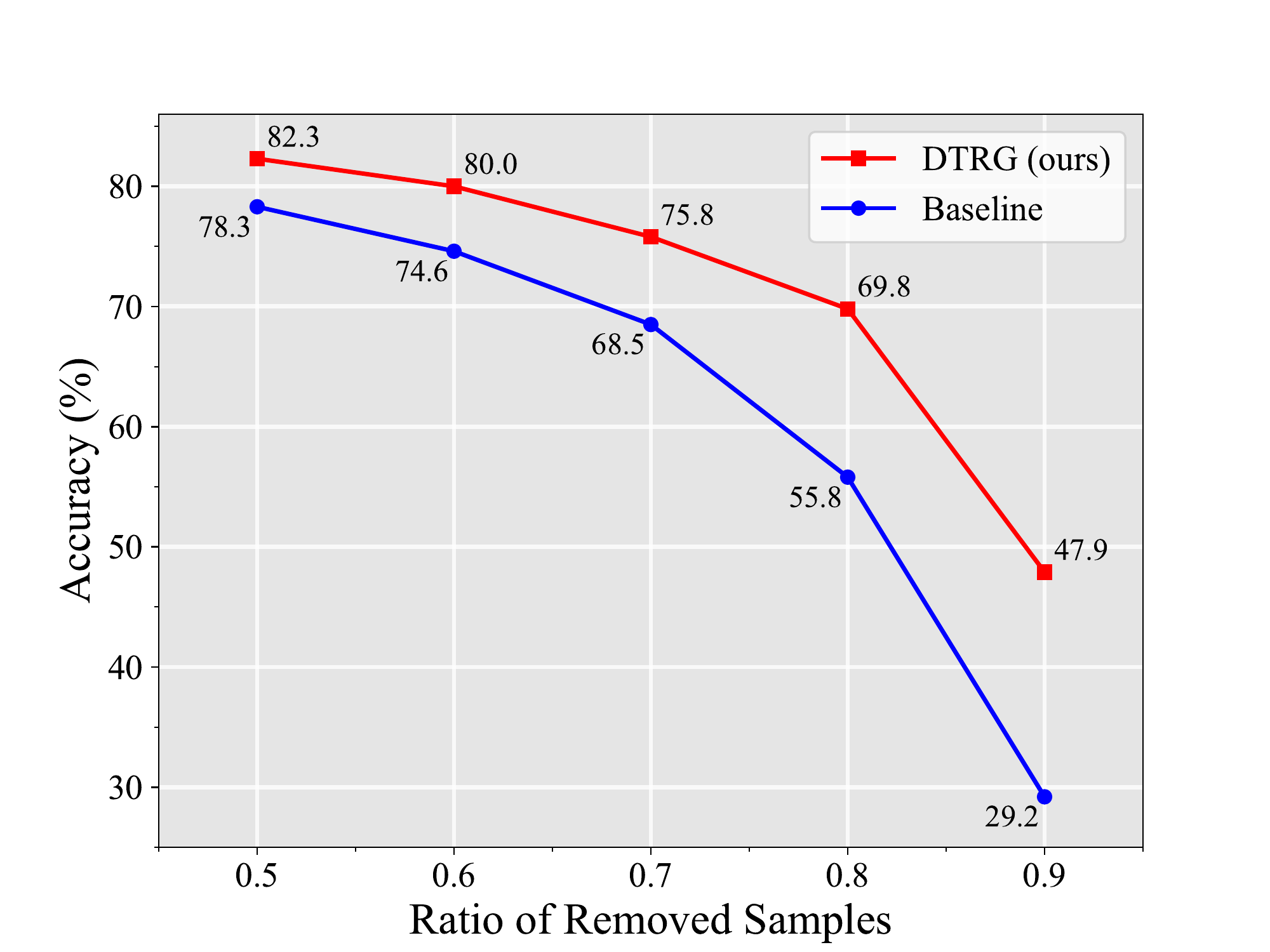}
    \caption{Classification performance under varying sample size of each class.}
    \label{fig:sparse_sampling}
\end{figure}
\begin{figure}[t]
    \centering
    \includegraphics[width=0.98\linewidth, trim=20 5 20 45,clip]{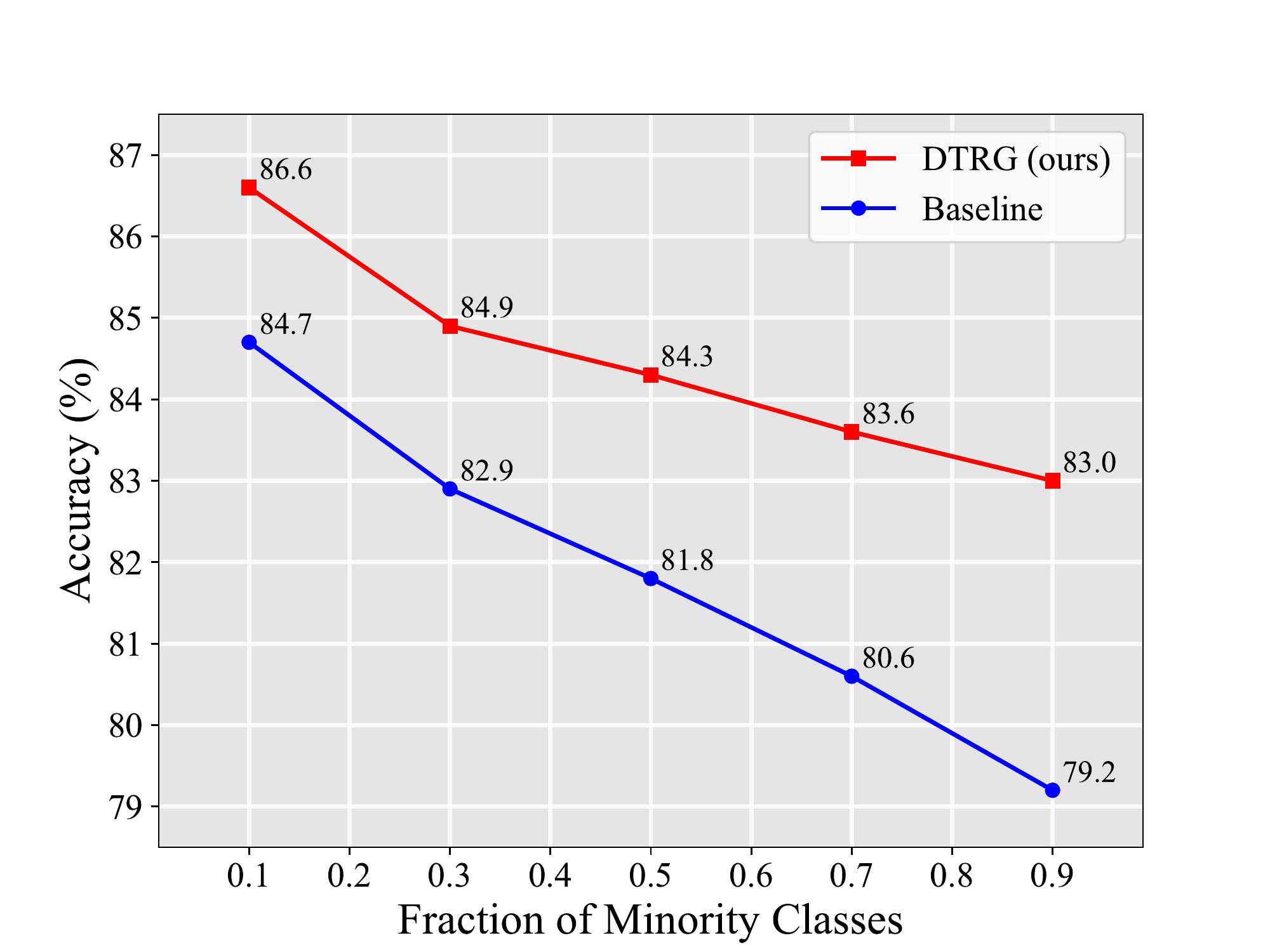}
    \caption{Evaluation with a long tailed data distribution.}
    \label{fig:imbalance_data}
\end{figure}

\subsection{Robust Against Sparse and Imbalanced Data}\label{sec:sparse-imbalance}
In order to verify the robustness of our method against sparse and imbalanced data distributions, we conduct two more experiments on the CUB-200-2011 dataset with the ResNet50 backbone. 
As Table \ref{table:datasets} shown, the CUB-200-2011 dataset contains total 200 classes, where each class has a  relative balanced distribution (about 30 training samples for each class), which thus requires to adapt the CUB dataset to a more challenging sparse and imbalanced setting.

For evaluation on sparse data, we set the sparsity of the training set with a limited size of samples for each class (removing 50\% to 90\% of training samples), whereas the whole validation set is adopted in our experiments. 
As shown in Fig. \ref{fig:sparse_sampling}, our DTRG method shows superior robustness against sparse data, which is credited to utilize structural correlation in the target space, compared to its baseline competitor. 
\textcolor{black}{Specifically, as mentioned in the introduction, representation learning for sparse classes can be supported by samples of other correlated semantic classes, as information across classes can be shared via latent target relation graph regularization.}
Furthermore, the more sparse data, the larger performance gap gained by our model over the baseline.

For evaluation on class imbalance, we generate the training set by following the \textit{step imbalance} setting in \cite{Buda2018ASS} with an equal number of samples in minority and majority classes, respectively.
To construct an imbalanced training set, we randomly choose one half of the samples from the candidate minority classes, while keep original sample size for the majority classes. 
The magnitude of class imbalance by controlling the fraction of minority classes is varying from 0.1 to 0.9 in the experiment. 
We again use the whole validation set for testing. 
The results are shown in Fig. \ref{fig:imbalance_data}. 
It is observed that our method can always outperform the baseline with different magnitude of class imbalance.

\textcolor{black}{
As the iNaturalist 2018 dataset also exhibits enormous long-tailed imbalance, the results presented in Table \ref{table:inaturalist} can 
also support better robustness of the proposed DTRG method than its competing baseline as well as existing algorithms.
Similarly, our explanation of robust performance gap between our method and baseline can be credited by inter-class information transfer, imposed by our target relation graph regularization. 
}

\subsection{Ablation Studies}\label{subsec: ablation}
In the total loss function in \textbf{Eq.} (\ref{eq:total loss}), there are three loss terms -- the CE loss in \textbf{Eq.} (\ref{eq:celoss}) for typical supervision on classification, the Online Center loss in \textbf{Eq.} (\ref{eq:centerloss}) for minimizing the intra-class feature variations, and the Graph Similarity loss in \textbf{Eq.} (\ref{eq:euclidean}) for incorporating inter-class relations. 
In this section, the effects of modules in our proposed method with different hyper parameters are investigated and evaluated.
Then, effects of error gradients from extra supervision on middle layers are evaluated.
Finally, the inter-class relation augmentation is compared on three backbone networks to demonstrate its effectiveness.

\begin{table}[t]
\centering
\caption{Ablation studies on CUB,  CAR and AIR datasets with ResNet50,  ResNet101, DenseNet161, \textbf{ResNet50+mid}, \textbf{ResNet101+mid} and \textbf{DenseNet161+mid} backbones, respectively.}
\label{table: backbone_ablation}
\begin{tabular}{@{}l|cc|ccc@{}}
\toprule
Backbone                      & OCL                                                   & GSL                                                             & {CUB (\%)} & {CAR (\%)} & {AIR (\%)}  \\ \midrule
\multirow{3}{*}{ResNet50}     &                                                       &                                                                 & 85.6          & 92.9          & 90.7          \\  
                              & $\surd$                                               &                                                                 & 86.6          & 93.6          & 91.5          \\ 
                              & $\surd$                                               & $\surd$                                                         & 87.3          & 94.3          & 92.0          \\ \midrule
\multirow{3}{*}{ResNet50+mid} &                                                       &                                                                 & 87.7          & 93.6          & 91.5          \\ 
                              & $\surd$                                               &                                                                 & 87.9          & 94.1          & 92.3          \\ 
                              & $\surd$                                               & $\surd$                                                         & {88.3}        & {94.8}        & {93.0} \\ \midrule  \midrule

\multirow{3}{*}{ResNet101}       &                                                       &                                                                 & 86.4 	  & 93.2   & 91.4              \\ 
                                 & $\surd$                                               &                                                               & 87.1 	  & 93.8   & 92.1                 \\ 
                                 & $\surd$                                               & $\surd$                                                       & 87.7       & 94.4  & 92.8  \\ \midrule
\multirow{3}{*}{ResNet101+mid} &                                                       &                                                                 & 87.9       & 94.0  & 92.2     \\ 
                                 & $\surd$                                               &                                                               & 88.1       & 94.2  & 92.6      \\ 
                                 & $\surd$                                               & $\surd$                                                       &88.6        & 94.9    & 93.1       \\ \midrule  \midrule                           

\multirow{3}{*}{DenseNet161}     &                                                       &                                                                 & 87.3     & 93.3     & 92.4     \\ 
                                 & $\surd$                                               &                                                                 & 88.2     & 94.0     & 92.9     \\ 
                                 & $\surd$                                               & $\surd$                                                         & 88.4     & 94.8     & 93.5     \\ \midrule
\multirow{3}{*}{DenseNet161+mid} &                                                       &                                                                 & 88.4     & 93.6     & 92.7     \\ 
                                 & $\surd$                                               &                                                                 & \textbf{89.0}     & 94.2     & 93.1     \\ 
                                 & $\surd$                                               & $\surd$                                                         & \textbf{89.0}    & \textbf{94.8} & \textbf{94.0} \\ \bottomrule
\end{tabular}
\end{table}

\subsubsection{Impact of the Proposed Components}

\begin{figure*}[htbp]
    \centering
    \subfigure[Impact of $\beta$]{ 
            \includegraphics[width=0.31\linewidth, trim=5 0 60 50,clip]{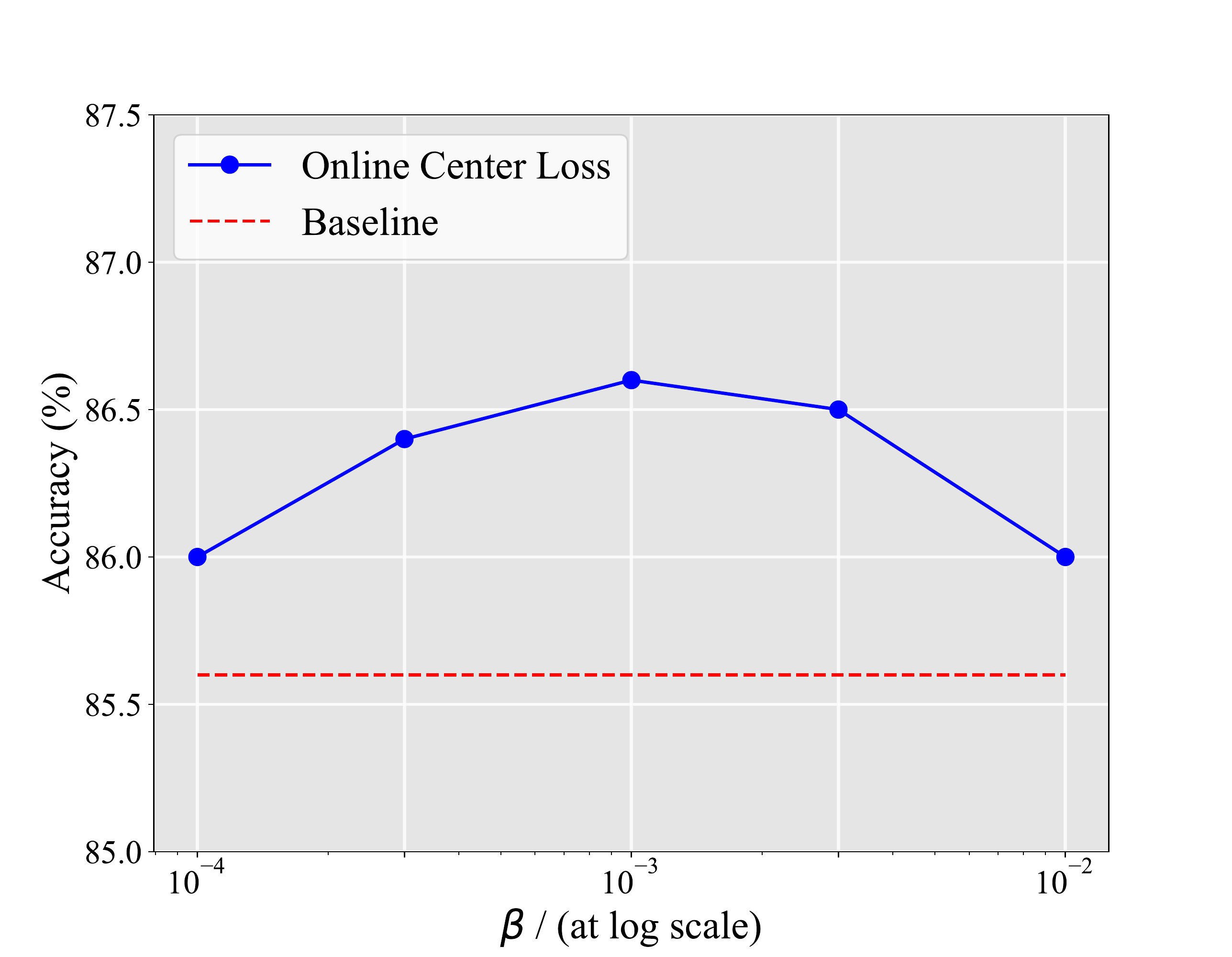}
            \label{fig:beta}}
    \subfigure[Impact of $\eta_\text{eu}$]{
            \includegraphics[width=0.31\linewidth, trim=5 0 60 50,clip]{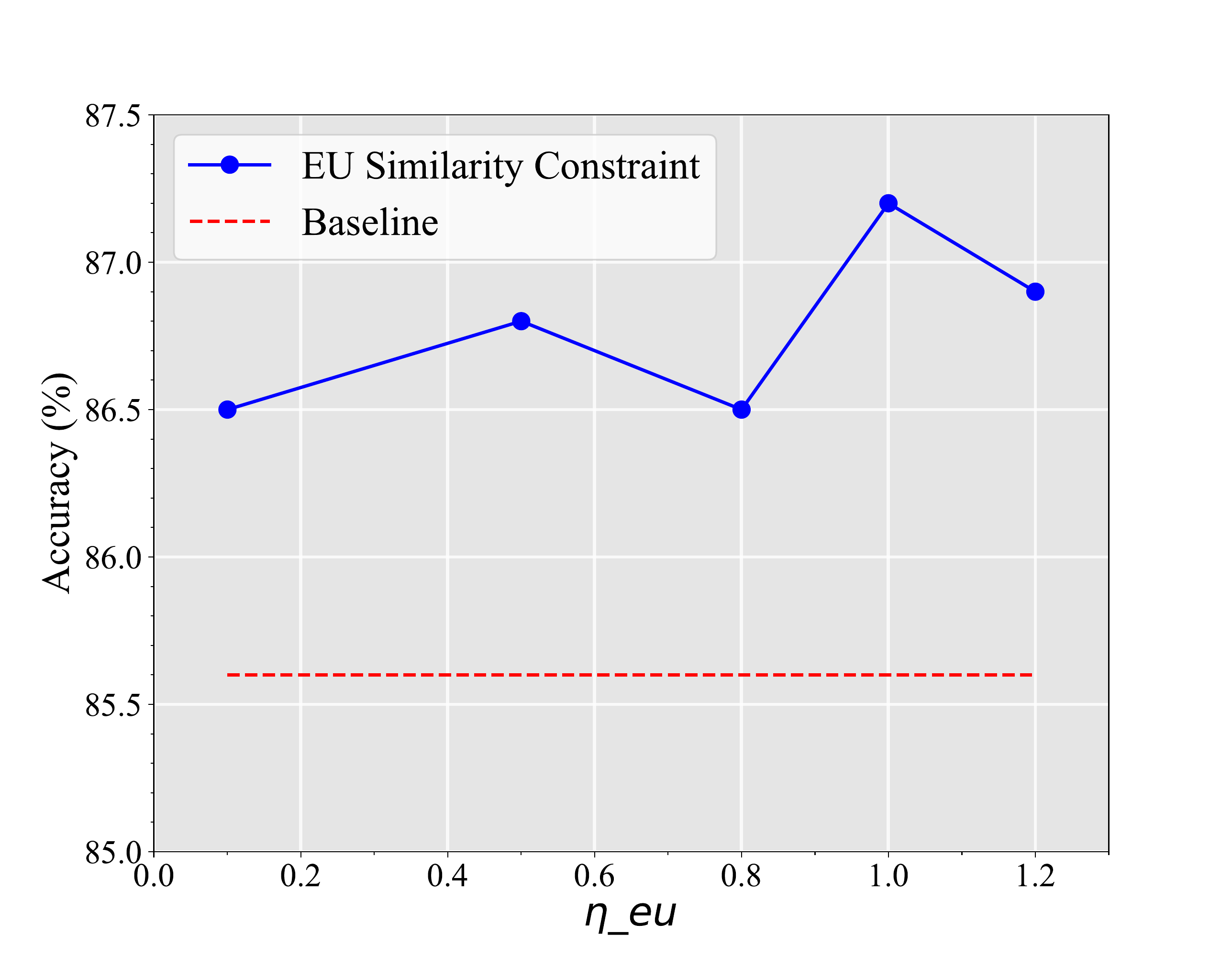}
            \label{fig:eta_eu}}
    \subfigure[Impact of $\eta_\text{kl}$]{
            \includegraphics[width=0.31\linewidth, trim=5 0 60 50,clip]{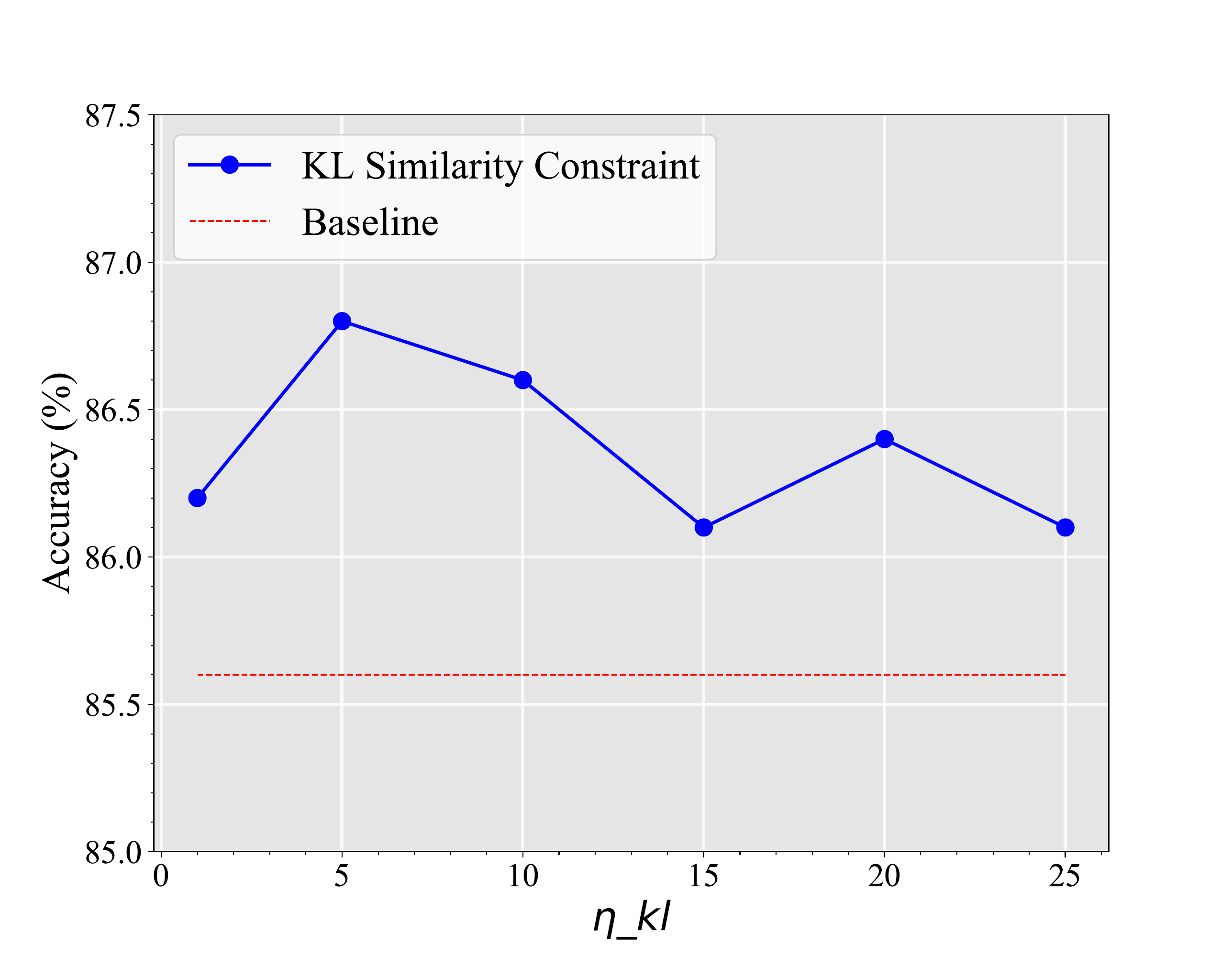}
            \label{fig:eta_kl}}
    \caption{Effects of hyper parameters for non-parametric online center loss and graph similarity loss with squared Euclidean distance or KL Divergence.}
    \label{fig:parameter}
\end{figure*}

We conduct ablation studies to demonstrate the effectiveness of the two key components in our model, including the non-parametric online center loss and the graph similarity constraint. 
Ablation studies are conducted on six different baselines, which adopt three different backbone networks, the ResNet50, the ResNet101 and the DenseNet161 with or without using mid-level branch on three different fine-grained datasets. 
From Table \ref{table: backbone_ablation}, we can find out that the effectiveness of the proposed modules with different backbones on multiple datasets are verified.
Especially, our models based on \textbf{backbone+mid} can always get the better performance than those based on vanilla backbone, which indicates that additionally combining features from mid-level can further improve discrimination of features. 
What's more, the deeper the network structure, the better the performance. 
With the \textbf{DenseNet161+mid} backbone, we can achieve the state-of-the-art performance on all three datasets.
As depicted in \textbf{Eq.} (\ref{eq:total loss}),  two hyper parameters $\beta$ and $\eta$ for non-parametric online center loss and graph similarity loss need to be tuned respectively. 
Note that we have proposed two options of graph similarity constraint, and hence we replace the hyper parameter $\eta$ with $\eta_\text{eu}$ and $\eta_\text{kl}$ for the two graph similarity constraints, corresponding to the squared Euclidean distance and the KL Divergence. 

\vspace{0.1cm}\noindent \textbf{Effects of hyper parameter $\beta$ --} In the first experiment, we add our non-parametric online center loss to the baseline as the only regularization, and vary $\beta$ from $1e-4$ to $1e-2$. The experiment is conducted on the CUB-200-2011 dataset with the ResNet50 backbone.
The results shown in Fig. \ref{fig:beta}, where the performance can generally be improved by adding our non-parametric online center loss. 
We can also observe that when $\beta= 1e-3$, the best result is gained, which can outperform the baseline by 1.01\% on classification accuracy. 
As a result, $\beta= 1e-3$ is fixed for other experiments.

\vspace{0.1cm}\noindent \textbf{Effects of hyper parameter $\eta$ --} To explore the influence of different similarity constraints, we set up two groups of comparative experiments for $\eta_\text{eu}$ and $\eta_\text{kl}$, respectively. 
Considering different properties, the parameter scales of $\eta_\text{eu}$ and $\eta_\text{kl}$ are different.
That is, $\eta_\text{eu}$ can be varied from 0.1 to 1.2 for similarity constraint based on the squared Euclidean distance, and $\eta_\text{kl}$ is set from 1 to 25 with an approximately equal interval for similarity constraint method based KL Divergence.
Experiments are conducted by adding our graph similarity loss as the only regularization with the above settings of parameter. 
The results are visualized in Fig. \ref{fig:eta_eu} and~\ref{fig:eta_kl}, which can be evidently observed that both similarity constraints can outperform baseline consistently.
Moreover, it is concluded that graph similarity constraint based on the squared Euclidean distance can perform slightly better than that based on the KL Divergence. 
The best result is achieved when we adopt the squared Euclidean distance with $\eta_\text{eu}=1$, which outperforms the baseline by 1.55\%. 
Consequently, the squared Euclidean distance is employed in our graph similarity constraint with $\eta=1$ in other  experiments. 

\vspace{0.1cm}\noindent
\textcolor{black}{ \textbf{Effects of starting epoch $T_{th}$ --} To explore the effects of different start epoch on classification performance when using our DTRG method, we set up a group of experiments with varying $T_{th}$ from 2 to 50 on the CUB-200-2011 dataset. 
Table \ref{table:epoch-ablation} shows that the best results can be achieved when the start epoch $T_{th}$ is set to be a small value (\eg 2 and 3),
which can be credited to good yet diverse initial feature centers of target relation graph by the pre-trained models.
Positive effects of feature regularization can be reduced with $T_{th}$ increasing, due to lack of exploration of latent target relation with convergence of feature centers.  
However, with any $T_{th}$, our DTRG can consistently outperform its baseline, which can again verify the superiority of our method.
}

\begin{table}[t]
\centering
\setlength{\tabcolsep}{2mm}
\caption{Effects of hyper-parameter $T_{th}$ for training with our DTRG method on the CUB dataset.}
\label{table:epoch-ablation}
\begin{tabular}{@{}l|ccccccc|c@{}}
\toprule
$T_{th}$ & 2    & 3    & 4    & 6    & 10   & 20   & 50  & Baseline   \\ \midrule
Acc (\%)   & 87.3 & 87.3 & 87.2 & 87.0 & 86.9 & 86.4 & 86.3 & 85.6 \\ \bottomrule
\end{tabular}
\end{table}

\begin{figure*}[thb]
\centering
\subfigure[Baseline]{
        \includegraphics[width=0.3\linewidth, trim=70 70 70 70 ,clip]{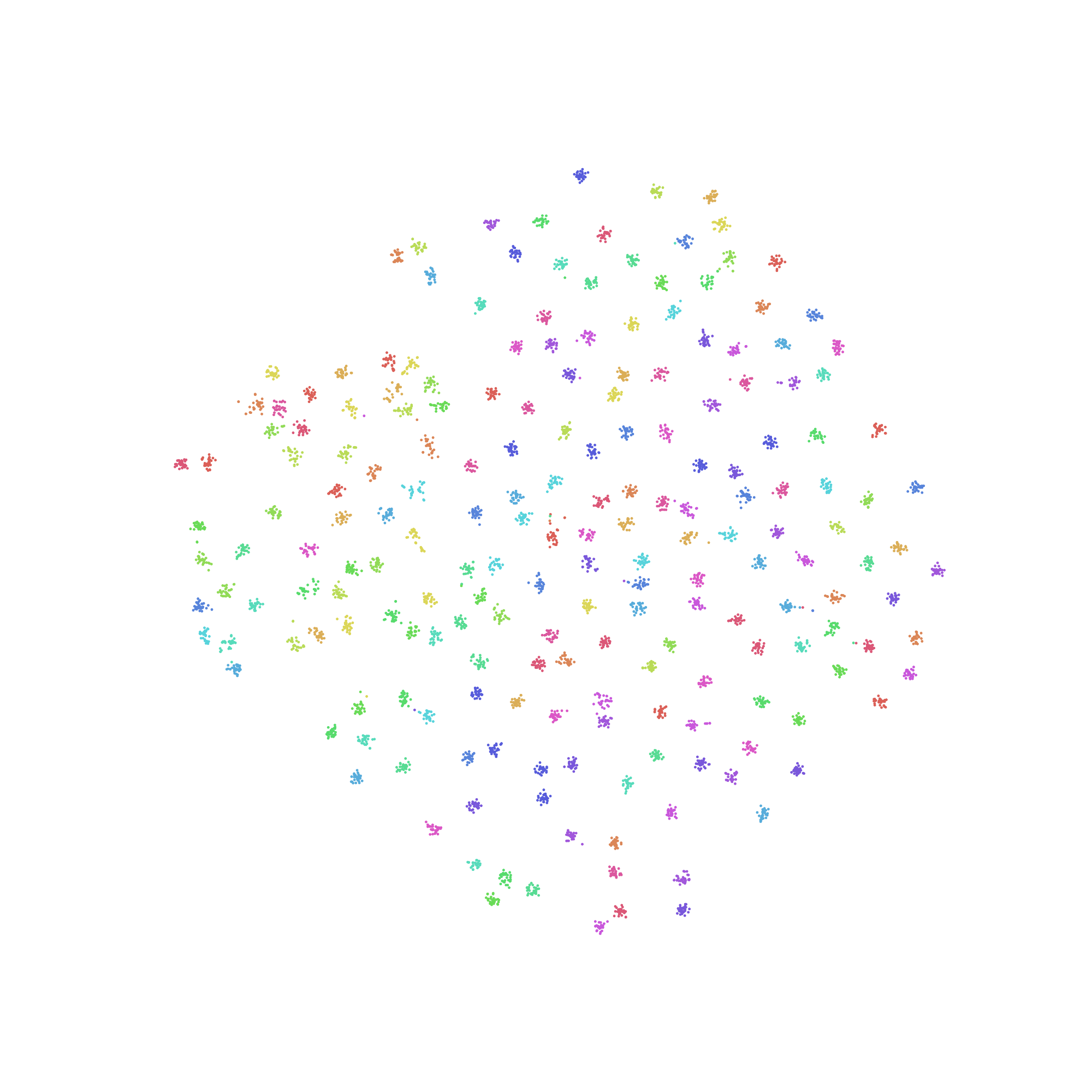}
\label{fig:subfigure1}}
\quad
\subfigure[the OCL (Ours)]{
        \includegraphics[width=0.3\linewidth,trim=70 70 70 70 ,clip]{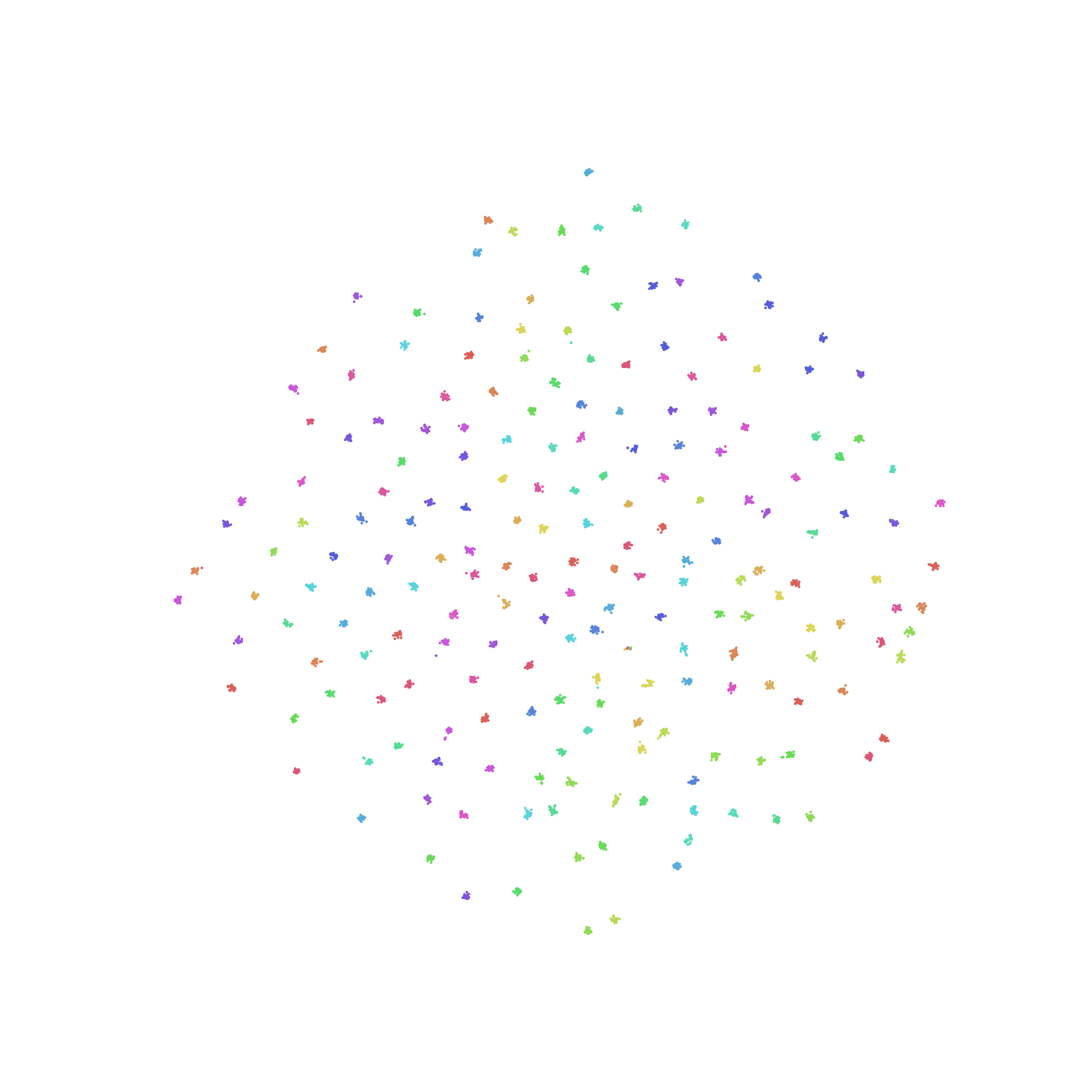}
\label{fig:subfigure2}}
\quad
\subfigure[the GSL (Ours)]{
        \includegraphics[width=0.3\linewidth, trim=70 70 70 70 ,clip]{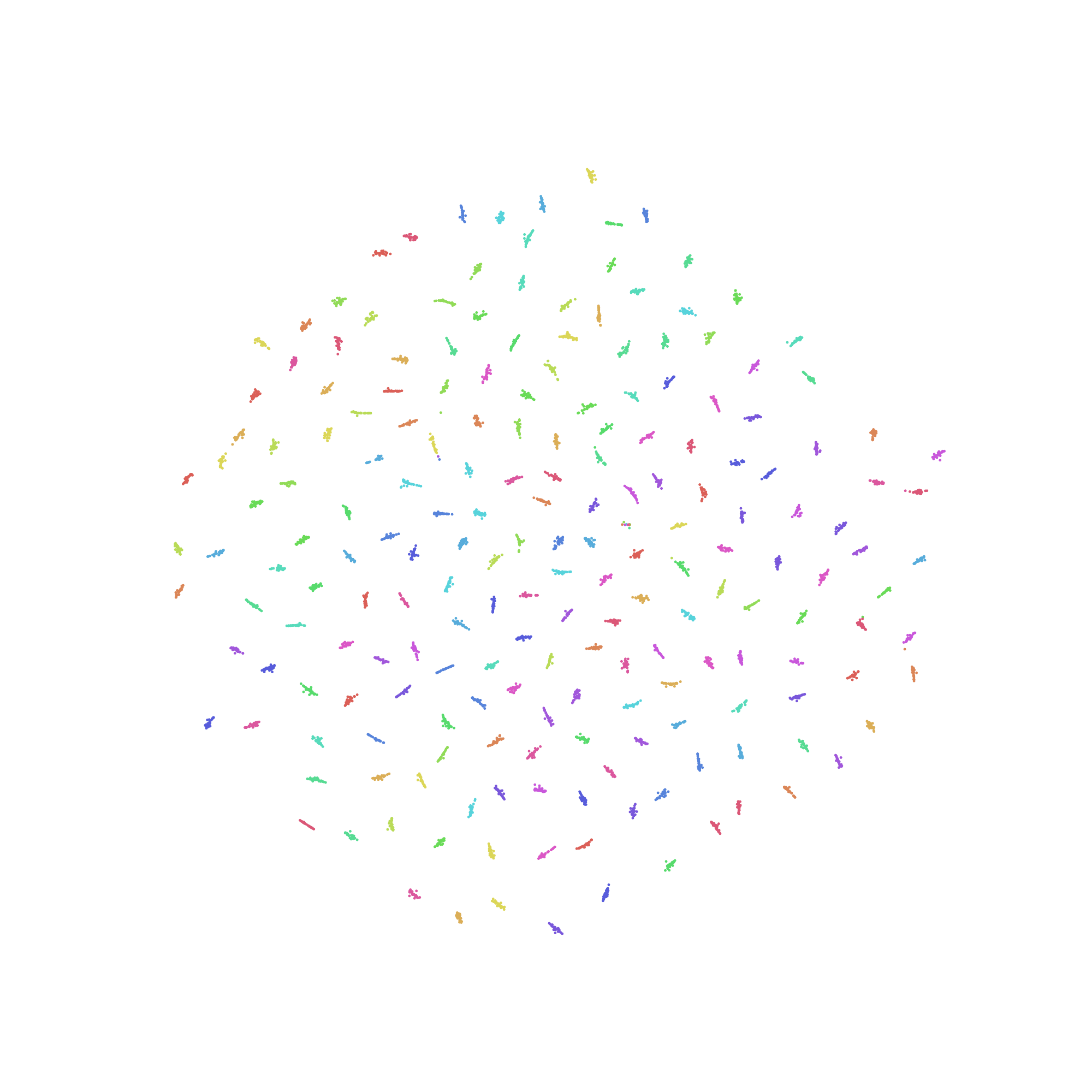}
\label{fig:subfigure3}}

\caption{Visualization for baseline, our non-parametric online center loss (OCL) and graph similarity loss (GSL) on the CUB-200-2011 training dataset with the t-SNE \cite{Maaten2008VisualizingDU}. The features for visualization are from the same last block of the ResNet-50 backbone of all three models. (Best viewed in color)}
\label{fig:visualization}
\end{figure*}

\subsubsection{Effects of Error Gradients from Extra Supervision on Middle Layers}
Inspired by existing works \cite{huang2021snapmix}, we add another classification branch in the middle of our backbone network to boost classification performance. 
Since generality of the proposed DTRG scheme, our method can also be applied to regularize mid-level feature encoding with self-supervised structural relation in target space.
We choose the \textbf{ResNet50+mid} as our backbone and conduct experiments on the CUB-200-2011 dataset.
Ablation studies on the online class loss and the geometric similarity loss are shown in the Table \ref{table:detach}, where \textbf{w/o detach} and  \textbf{w/ detach} indicate whether gradients of errors generated from supervision on the mid-level layers will be propagated backward to the backbone networks or not. 
In other words, the models \textbf{w/ detach} can be considered as an ensemble of classifiers on feature encoding with different layers, while the models \textbf{w/o detach} use extra supervision on mid-level feature encoding layers. 
We can conclude that the models \textbf{w/ detach} consistently gain the better performance than those with gradient back-propagation (\ie models \textbf{w/o detach}). 
Our explanation is that a $Max~Pooling$ Layer is employed in the extra branch on middle layers, which could disturb gradients backward from the $Average~Pooling$ in the main branch. 
In the remaining experiments, we all use the models 
without gradients backward from mid-layers to the main backbone. 

\begin{table}[t]
\centering
\caption{Effects for error gradients from mid-level layers backward to the backbone.
Experiments are conducted on CUB-200-2011 (CUB) dataset with \textbf{ResNet50+mid} backbone.}
\label{table:detach}
\begin{tabular}{@{}lcccc@{}}
\toprule
\multirow{2}{*}{Baseline}     & \multirow{2}{*}{OCL}      & \multirow{2}{*}{GSL}                 & \multicolumn{2}{c}{{CUB} (\%)} \\ \cmidrule(l){4-5} 
                              &                           &                                      & w/o detach   & w/ detach  \\ \midrule
\multirow{3}{*}{ResNet50+mid} &                           &                                      & 86.7         & 87.7       \\
                              & $\surd$                   &                                      & 87.0         & 87.9       \\
                              & $\surd$                   & $\surd$                              & 87.1         & 88.3       \\ \bottomrule
\end{tabular}
\end{table}

\begin{table}[t]
\centering
\caption{\textcolor{black}{Evaluation on the inter-class relation augmentation on the CUB, the CAR and the AIR datasets. Experiments are implemented with \textbf{ResNet50+mid}, the \textbf{ResNet101+mid} and \textbf{DenseNet161+mid} backbones, respectively.}}
\label{table:mixup-snapmix}
\setlength{\tabcolsep}{1mm}
\begin{tabular}{@{}l|ccc|ccc@{}}
\toprule
Backbone                         & Mixup & CutMix  & SnapMix & {CUB (\%)} & {CAR (\%)} & {AIR (\%)} \\ \midrule
\multirow{4}{*}{\begin{tabular}[c]{@{}l@{}}ResNet50 \\ +mid\end{tabular}}    &         &         &         & 88.3    & 94.8    & 93.0    \\
                                 & $\surd$ &         &         & 88.4    & 95.2    & 93.5    \\
                                 &         &$\surd$  &         & 88.3    & 95.2    & 93.0    \\
                                 &         &         & $\surd$ & 88.8    & 95.2    & 94.1    \\ \midrule \midrule
\multirow{4}{*}{\begin{tabular}[c]{@{}l@{}}ResNet101 \\ +mid\end{tabular}}   &         &         &         & 88.6    & 94.9    & 93.1    \\ 	 	 
                                 & $\surd$ &         &         & 88.6    & 95.4    & 93.6    \\
                                 &         &$\surd$  &         & 89.2    & 95.1    & 93.2    \\
                                 &         &         & $\surd$ & 89.7    & 95.3    & 94.2    \\ \midrule \midrule
\multirow{4}{*}{\begin{tabular}[c]{@{}l@{}}DenseNet161 \\ +mid\end{tabular}} &         &         &         & 89.0    & 94.8    & 94.0    \\
                                 & $\surd$ &         &         & 89.7    & 95.4    & 94.1    \\
                                 &         &$\surd$  &         & 89.5    & 95.4    & 93.8    \\
                                 &         &         & $\surd$ & \textbf{90.1} & \textbf{95.7} & \textbf{94.8} \\ \bottomrule
\end{tabular}
\end{table}

\subsubsection{Evaluation on Inter-class Relation Augmentation}
One more experiment is conducted to verify the effectiveness of our proposed inter-class relation augmentation on the CUB-200-2011, the Stanford Cars, and the FGVC-Aircraft datasets with the backbone of the \textbf{ResNet50+mid}, the \textbf{ResNet101+mid} and the \textbf{DenseNet161+mid}. 
We explore three types of data augmentation methods (\ie the Mixup, the CutMix and the SnapMix) as explained in Sec. \ref{subsec: relation}, which share the identical structure with the only difference lying on hyper parameter $\alpha$ for weighting the proportion of two images. 
\textcolor{black}{As shown in Table \ref{table:mixup-snapmix}, 
the first row of each block is our method based on different backbone without using inter-class relation augmentation, while the remaining three are enhanced with inter-class relation augmentation based on the Mixup, the CutMix and the SnapMix, respectively. 
It is observed that our proposed inter-class relation augmentation can further improve classification performance on all datasets with different backbones.
Moreover, our DTRG with the SnapMix based augmentation can perform slightly better than the ones using the Mixup and the CutMix, owing to its usage of the more precise semantic-relatedness proportion (\ie $\lambda$ in \textbf{Eq.} (\ref{eq:mixup1})) of interpolated images.
}

\subsection{Visualization}

To illustrate the rationale of our proposed non-parametric online center loss and graph similarity constraint modules, we utilize the t-SNE \cite{Maaten2008VisualizingDU} to project the features from the last block of trained models (\eg models based on the ResNet-50 backbone in Fig. \ref{fig:framework}(a)) into 2d space.
In our method, feature centers of each category are generated by the mean of features of samples from the same category, which can be viewed as a simple yet robust prototype of all samples within each class.
Intra-class similarity in feature space can be achieved by minimizing average distance to feature centers from all data points of each category.
What’s more, the proposed target relation graph between different feature centers can serve as additional supervision signals to incorporate inter-class relation into representation learning, which thus can produce a structurally distributed feature space with limited principle directions (affected by highly-correlated neighbouring classes).
As shown in Fig. \ref{fig:subfigure2}, visual representations of each class will be more compact via the proposed online center loss (\ie the OCL), while with the constraint of our Target Relation Graph (\ie the GSL), intra-class samples can be more structurally distributed along a principle direction, as shown in Fig. \ref{fig:subfigure3}. 
When using our method, the feature centers of each class are updated in an iterative manner, which will affect generation of target relation graph.
Therefore, inter-class correlation can be dynamically varying during model training.
However, our DTRG method on enforcing the model to discover visual patterns across classes can gain better representations. 

\begin{figure}[t]
    \centering
    \includegraphics[width=\linewidth]{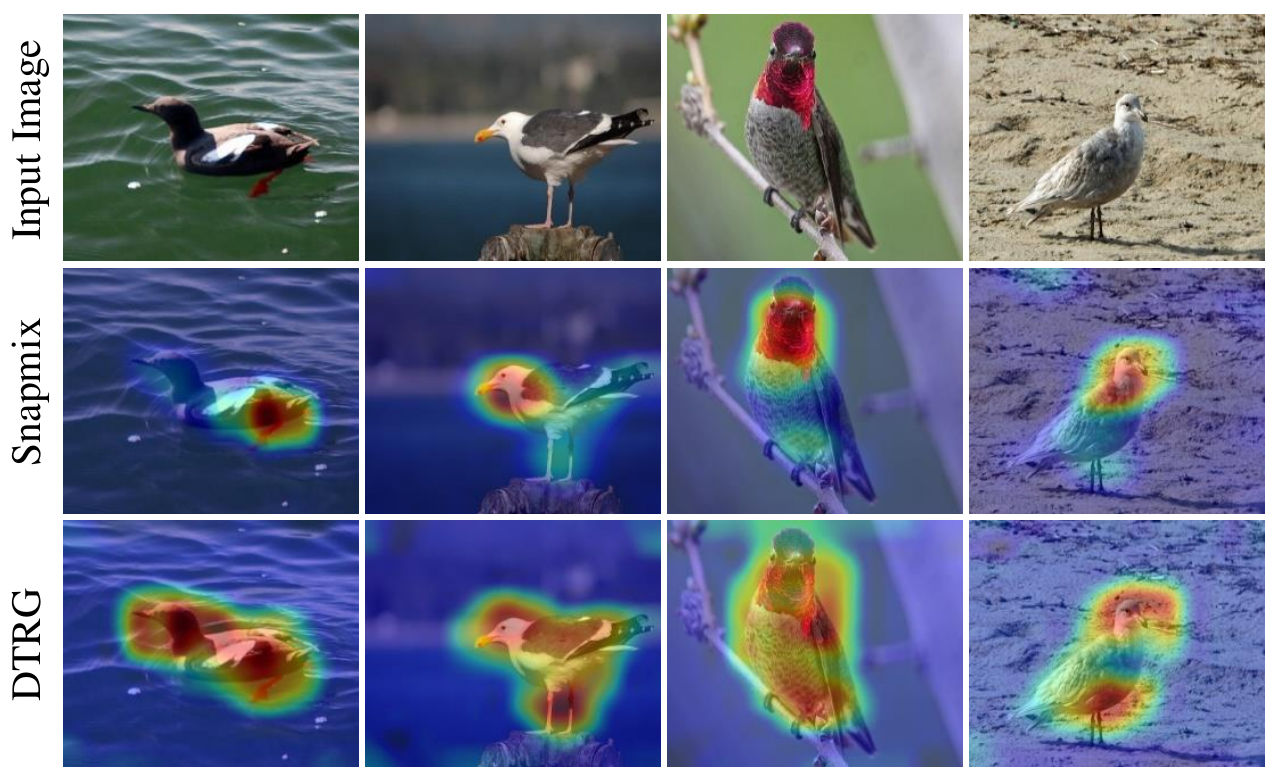}
    \caption{\textcolor{black}{Comparison of the Gradient CAM on images by the SnapMix and our DTRG. The first row shows vanilla input images, the second row shows the Gradient CAMs by the SnapMix, and the third row shows the Gradient CAMs by our DTRG. 
Evidently, discriminative information in object representations learned by the SnapMix is from local regions, while our DTRG can enforce the model to focus on not only discriminative regions but also less informative parts of objects guided by inter-class correlation.    
    (Best viewed in color)
    }}    
    \label{fig:grad_cam}
\end{figure} 

To show visual patterns captured by the proposed DTRG, we compare the attention maps of our DTRG with the SnapMix\cite{huang2021snapmix} method on the CUB-200-2011 dataset. 
Visualization based on the Gradient CAMs \cite{Selvaraju2019GradCAMVE} is displayed in Fig. \ref{fig:grad_cam}.
It is observed that the SnapMix method depends on features extracted from limited discriminative regions, while our DTRG can enforce the model to  focus on not only discriminative regions, but also less informative parts of objects guided by inter-class semantic correlation.
Visualized results demonstrate that our method can improve representations by capturing discriminative information from less informative regions.
Consequently, it is proven that our DTRG method can encourage more discriminative features.

\section{Conclusion}
\label{sec: conclusion}
In this paper, we propose a self-supervised feature regularization method on deep representation learning for  
visual classification, which can incorporate
inter-class correlation into features and also mitigate intra-class feature variations.
Superior performance can further be achieved via the proposed inter-class relation augmentation.
Extensive experiments on popular benchmarks of multiple visual classification tasks
can demonstrate the effectiveness of each proposed modules, consistently achieving the state-of-the-art performance.
Moreover, our method can be more robust against sparse and long-tail data distributions owing to exploiting latent target relation.




\ifCLASSOPTIONcaptionsoff
  \newpage
\fi



\bibliographystyle{IEEEtran}
\bibliography{IEEEabrv,tip2021}
%



%


\begin{IEEEbiography}[{\includegraphics[width=1in,height=1.25in,clip,keepaspectratio]{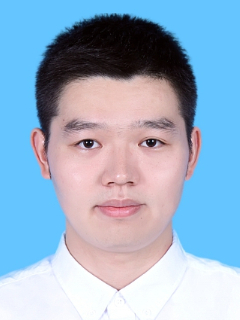}}]{Kangjun Liu}
received the B.E. degree in the College of Mechanical and Vehicle Engineering from Hunan University, China, in 2016. He is currently pursuing the Ph.D. degree in the Shien-Ming Wu School of Intelligent Engineering, South China University of Technology. His research interests are in computer vision, deep learning and pattern recognition.
\end{IEEEbiography}

\begin{IEEEbiography}[{\includegraphics[width=1in,height=1.25in,clip,keepaspectratio]{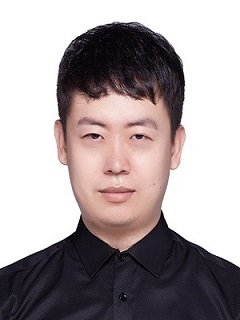}}]{Ke Chen} is currently an Associate Professor with the School of Electronic and Information Engineering, South China University of Technology (SCUT). 
Before joining the SCUT, he was a Postdoctoral Research Fellow with the Department of Signal Processing, Tampere University of Technology, Finland.
He received the B.E. degree in automation and the M.E. degree in software engineering from Sun Yat-sen University in 2007 and 2009, respectively, and the Ph.D. degree in computer vision from Queen Mary University of London in 2013. 
His research interests include computer vision, pattern recognition, neural dynamic modeling, and robotic inverse kinematics.
\end{IEEEbiography}

\begin{IEEEbiography}[{\includegraphics[width=1in,height=1.25in,clip,keepaspectratio]{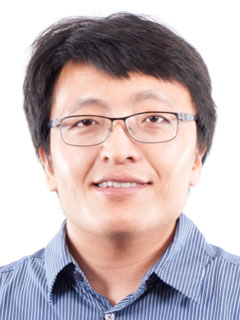}}]{Kui Jia}
received the B.E. degree from North-western Polytechnic University, Xi’an, China, in 2001, the M.E. degree from the National University of Singapore, Singapore, in 2004, and the Ph.D. degree in computer science from the Queen Mary University of London, London, U.K., in 2007. He was with the Shenzhen Institute of Advanced Technology of the Chinese Academy of Sciences, Shenzhen, China, Chinese University of Hong Kong, Hong Kong, the Institute of Advanced Studies, University of Illinois at Urbana-Champaign, Champaign, IL, USA, and the University of Macau, Macau, China. He is currently a Professor with the School of Electronic and Information Engineering, South China University of Technology, Guangzhou, China. His recent research focuses on theoretical deep learning and its applications in vision and robotic problems, including deep learning of 3D data and deep transfer learning.
\end{IEEEbiography}







\end{document}